%% file: main.tex
\documentclass[sigconf]{acmart}

\newif\ifsupponly
\supponlyfalse

\AtBeginDocument{%
  }

\copyrightyear{2026}
\acmYear{2026}
\setcopyright{cc}
\setcctype{by}
\acmConference[SA Conference Papers '26]{SIGGRAPH Asia 2026 Conference Papers}{December 01--04, 2026}{Kuala Lumpur, Malaysia}
\acmBooktitle{SIGGRAPH Asia 2026 Conference Papers (SA Conference Papers '26), December 01--04, 2026, Kuala Lumpur, Malaysia}
\acmDOI{10.1145/3829340.3842210}
\acmISBN{979-8-4007-2842-6/2026/12}

\usepackage{graphicx}
\usepackage{enumitem}
\usepackage{multirow}
\usepackage{pgfplots}
\pgfplotsset{compat=1.18}
\usepackage{xcolor}
\usepackage{natbib}
\usepackage{booktabs}

\renewcommand\dbltopfraction{0.92}
\renewcommand\textfraction{0.05}
\renewcommand\dblfloatpagefraction{0.85}

\ifsupponly
\input{supp_preamble}
\fi

\definecolor{first}{RGB}{220, 20, 60}     
\definecolor{second}{RGB}{255, 140, 0}    
\definecolor{third}{RGB}{34, 139, 34}     

\newcommand{\rev}[1]{#1}

\begin{document}

\ifsupponly

\title{Supplementary Material: MVID}
\maketitle

\appendix
\input{section/7_supp}

\clearpage

\bibliographystyle{ACM-Reference-Format}
\bibliography{references}

\else

\title{MVID: Feed-Forward Multi-View Intrinsic Image Decomposition}

\author{Kang Du}
\orcid{0009-0006-4182-9765}
\affiliation{%
  \institution{The Hong Kong University of Science and Technology (Guangzhou)}
  \city{Guangzhou}
  \country{China}}
\email{kdu800@connect.hkust-gz.edu.cn}

\author{Duotun Wang}
\orcid{0009-0005-4393-5230}
\affiliation{%
  \institution{The Hong Kong University of Science and Technology (Guangzhou)}
  \city{Guangzhou}
  \country{China}}
\email{dwang866@connect.hkust-gz.edu.cn}

\author{Wanling Li}
\orcid{0009-0008-7112-4410}
\affiliation{%
  \institution{The Hong Kong University of Science and Technology (Guangzhou)}
  \city{Guangzhou}
  \country{China}}
\email{wli608@connect.hkust-gz.edu.cn}

\author{Yirui Guan}
\orcid{0009-0006-0603-748X}
\affiliation{%
  \institution{Tencent}
  \city{Shenzhen}
  \country{China}}
\email{guan1rui@gmail.com}

\author{Zeyu Wang}
\authornote{Corresponding author.}
\orcid{0000-0001-5374-6330}
\affiliation{%
  \institution{The Hong Kong University of Science and Technology (Guangzhou)}
  \city{Guangzhou}
  \country{China}}
\affiliation{%
  \institution{The Hong Kong University of Science and Technology}
  \city{Hong Kong}
  \country{China}}
\email{zeyuwang@ust.hk}

\begin{abstract}
Intrinsic image decomposition aims to recover material and illumination factors from RGB observations, but real-world images entangle reflectance with illumination, visibility, shadows, and non-diffuse appearance.
Recent single-image methods address this entanglement with a residual image formation model, decomposing RGB into albedo, diffuse shading, and a non-diffuse residual.
However, applying such decomposition independently to consecutive frames lacks scene-level context, leading to inconsistent factor assignments, albedo drift, and leakage across views.
Meanwhile, multi-view inverse-rendering methods recover properties, such as material, lighting, and geometry, but they rely on synthetic data due to the highly uncertain estimation, and do not generalize to supervision by real images.
We present \textbf{MVID}, \textbf{M}ulti-\textbf{V}iew \textbf{I}ntrinsic image \textbf{D}ecomposition, a feed-forward framework built on the residual image formation model.
MVID builds a scene-level multi-view representation and decodes a view-consistent albedo together with coherent per-view shading and residual factors through a factor query adapter, while using the same image formation model for self-supervised RGB reconstruction on unlabeled real-world sequences.
Experiments on indoor, real-world, and outdoor benchmarks show that MVID improves both per-frame decomposition quality and cross-view consistency over single-view intrinsic, generative intrinsic, and multi-view inverse-rendering baselines.
The resulting view-stable factors support practical image-space applications, including multi-view consistent illumination editing and specularity removal.
The project page for this paper is at \url{https://github.com/dukang/MVID}.
\end{abstract}

\ccsdesc[500]{Computing methodologies~Rendering}
\ccsdesc[500]{Computing methodologies~Reconstruction}
\ccsdesc[500]{Computing methodologies~Computer vision representations}

\keywords{Intrinsic Image Decomposition, Multi-view Reconstruction, Feed-forward Scene Modeling}

\begin{teaserfigure}
\centering
\includegraphics[width=0.975\textwidth]{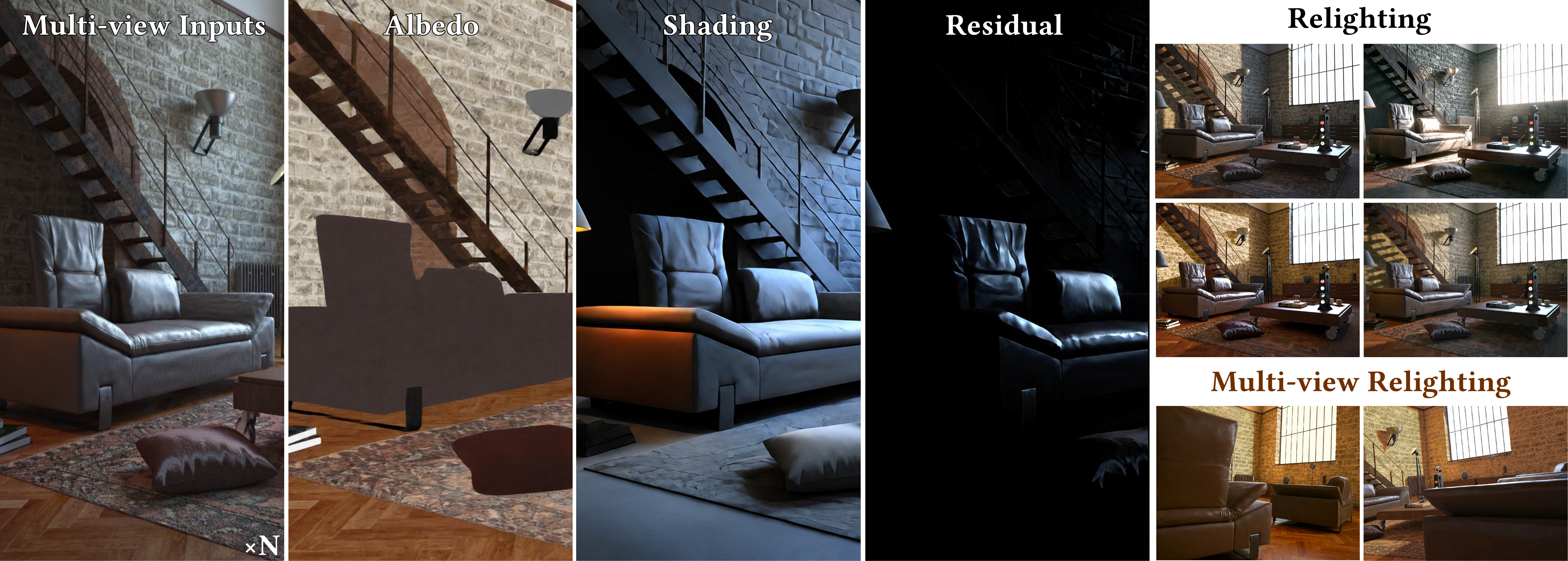}
\caption{
\textbf{Multi-view consistent intrinsic image decomposition and image-space illumination editing by MVID.}
Given multi-view RGB inputs (left), MVID decomposes each view into albedo, diffuse shading, and a non-diffuse residual.
Editing the diffuse shading or residual component while keeping albedo fixed enables controllable illumination appearance changes within a single view (top right) and consistent edits across views (bottom right).
}
\label{fig:teaser}
\end{teaserfigure}

\maketitle

\input{section/1_introduction}
\input{section/2_relatedwork}
\input{section/3_method2}
\input{section/4_experiments2}
\input{section/5_conclusion}

\begin{acks}
This work was supported by the Guangdong Basic and Applied Basic Research
Foundation (No.~2026A1515011138) and the Guangdong S\&T Program
HKUST-HKUST(GZ) ``1+1+1'' Joint Funding (G\_2025\_034).
\end{acks}

\bibliographystyle{ACM-Reference-Format}
\bibliography{references}

\fi

\end{document}

%% file: supp_preamble.tex
\usepackage{tcolorbox}
\usepackage{colortbl}
\usepackage{needspace}

%% file: section/7_supp.tex

\setcounter{figure}{0}
\setcounter{table}{0}
\renewcommand{\thefigure}{S\arabic{figure}}
\renewcommand{\thetable}{S\arabic{table}}

\renewcommand{\topfraction}{0.95}
\renewcommand{\bottomfraction}{0.7}
\renewcommand{\textfraction}{0.05}
\renewcommand{\floatpagefraction}{0.85}
\renewcommand{\dbltopfraction}{0.95}
\renewcommand{\dblfloatpagefraction}{0.85}
\setcounter{topnumber}{4}
\setcounter{dbltopnumber}{4}
\setcounter{totalnumber}{6}

\definecolor{mvidblue}{RGB}{30, 80, 140}
\definecolor{mvidlight}{RGB}{235, 240, 250}
\definecolor{mvidaccent}{RGB}{200, 90, 60}
\definecolor{mvidgray}{RGB}{120, 120, 120}
\definecolor{tabrowgray}{RGB}{245, 245, 245}

\newtcolorbox{takeaway}[1][]{
  colback=mvidlight,
  colframe=mvidblue,
  boxrule=0.6pt,
  arc=2pt,
  left=8pt,
  right=8pt,
  top=6pt,
  bottom=6pt,
  fonttitle=\bfseries\small\color{white},
  coltitle=white,
  title={Takeaway},
  #1
}

\newtcolorbox{notationbox}[1][]{
  colback=white, colframe=mvidgray,
  boxrule=0.4pt, arc=1pt,
  left=8pt, right=8pt, top=6pt, bottom=6pt,
  fonttitle=\bfseries\small,
  title={Notation},
  #1
}

\newcommand{\sectionaccent}{%
  \noindent\textcolor{mvidblue}{\rule{\linewidth}{1pt}}%
  \par\vspace{-2pt}%
  \noindent\textcolor{mvidblue}{\rule{0.25\linewidth}{0.4pt}}%
  \par\vspace{4pt}%
}


\begin{tcolorbox}[
  colback=mvidlight,
  colframe=mvidblue,
  boxrule=0.6pt,
  arc=2pt,
  left=10pt, right=10pt, top=8pt, bottom=8pt,
  title={\textbf{Contents}},
  fonttitle=\color{white}\bfseries
]
\small
\begin{itemize}[leftmargin=1.3em, itemsep=2pt, topsep=2pt]
  \item[\textcolor{mvidblue}{\textbf{\ref{sec:supp_impl}.}}] Implementation and architecture details
  \item[\textcolor{mvidblue}{\textbf{\ref{sec:supp_training_data}.}}] Training data, datasets, and training details
  \item[\textcolor{mvidblue}{\textbf{\ref{sec:supp_eval}.}}] Evaluation protocols
  \item[\textcolor{mvidblue}{\textbf{\ref{sec:supp_extra_quant}.}}] Additional quantitative analyses
  \item[\textcolor{mvidblue}{\textbf{\ref{sec:supp_adapter_vis}.}}] Adapter visualization and factor-specific cue analysis
  \item[\textcolor{mvidblue}{\textbf{\ref{sec:supp_mvinverse}.}}] Discussion of concurrent work: MVInverse
  \item[\textcolor{mvidblue}{\textbf{\ref{sec:supp_qual}.}}] Additional qualitative results
  \item[\textcolor{mvidblue}{\textbf{\ref{sec:supp_apps}.}}] Application details
  \item[\textcolor{mvidblue}{\textbf{\ref{sec:supp_failures}.}}] Failure cases and limitations
\end{itemize}
\end{tcolorbox}

\vspace{4pt}

\begin{notationbox}
We follow the same symbols as the main paper. Input views are denoted $\{\mathbf{I}_i\}_{i=1}^{V}$. The hierarchy of cross-view aggregated tokens is $\{\mathbf{T}^{l}\}_{l=1}^{L_T}$ with $\mathbf{T}^{l}\!\in\!\mathbb{R}^{V\times P\times C}$. The flattened cross-view memory used by the adapter is $\bar{\mathbf{T}}^{l}\!\in\!\mathbb{R}^{(V\cdot P)\times C}$, and the factor-conditioned tokens for factor $k\!\in\!\{A,S,R\}$ are $\widetilde{\mathbf{T}}^{l}_{k}$. To avoid clashing with the loss symbol $\mathcal{L}$, we use $L_T$ for the number of aggregator layers (the main paper uses $L$).
\end{notationbox}

\Needspace*{6\baselineskip}
\section{Implementation and Architecture Details}
\label{sec:supp_impl}
\sectionaccent

\Needspace*{5\baselineskip}
\subsection{Multi-view Aggregator}
\label{sec:supp_arch_backbone}

We instantiate the multi-view aggregator with a VGGT-style transformer~\cite{Wang2025VGGT} that alternates frame-wise self-attention within a single view and global self-attention across all views. The encoder is initialized from a VGGT-style multi-view geometry checkpoint. We freeze it during the first 200 epochs (Sec.~\ref{sec:supp_training}) and fine-tune it jointly with the rest of the model afterwards.

The hierarchy $\{\mathbf{T}^{l}\}_{l=1}^{L_T}$ contains $L_T\!=\!24$ layers. Following DPT, the dense decoders read from four DPT-aligned layers; the multi-layer factor-query adapter (Sec.~\ref{sec:supp_arch_adapter}) is applied at these four layers ($l\!\in\!\rev{\{4,11,17,23\}}$), which match the feature levels consumed by the dense heads.

\Needspace*{5\baselineskip}
\subsection{Multi-Layer Factor Query Adapter}
\label{sec:supp_arch_adapter}

For each factor $k\!\in\!\{A,S,R\}$ we maintain learnable factor queries $\mathbf{Q}_{k}\!\in\!\mathbb{R}^{M\times C}$ with $M\!=\!8$ and $C\!=\!1024$. At each adapter layer $l$, we perform two cross-attentions: a \emph{read} from the flattened cross-view memory and a \emph{write} back to the per-view dense tokens.

\paragraph{Read.}
\begin{equation}
\mathbf{Z}^{l}_{k}=\mathrm{CrossAttn}\!\left(\mathbf{Q}_{k},\bar{\mathbf{T}}^{l}\right),\qquad \mathbf{Z}^{l}_{k}\!\in\!\mathbb{R}^{M\times C}.
\end{equation}
Flattening across views before the read lets each query gather scene-level evidence rather than within-frame evidence only.

\paragraph{Write.}
\begin{equation}
\widetilde{\mathbf{T}}^{l}_{k}=(1-\alpha^{l}_{k})\,\mathbf{T}^{l}+\alpha^{l}_{k}\,\mathrm{CrossAttn}\!\left(\mathbf{T}^{l},\mathbf{Z}^{l}_{k}\right).
\end{equation}
This is algebraically equivalent to the residual form used in the main paper:
\[
\widetilde{\mathbf{T}}^{l}_{k}
=
\mathbf{T}^{l}
+
\alpha^{l}_{k}
\big(
\mathrm{CrossAttn}(\mathbf{T}^{l},\mathbf{Z}^{l}_{k})
-
\mathbf{T}^{l}
\big).
\]
Special (non-patch) tokens are passed through unchanged.
The gate $\alpha^{l}_{k}\!\in\!\mathbb{R}$ is a \emph{scalar} (per-factor, per-layer), initialized to $0.1$ and bounded to $[0,1]$ via sigmoid reparameterization. We use scalar gates for simplicity and efficiency.

\paragraph{Difference from FiLM.}
FiLM~\cite{perez2018film} produces channel-wise affine modulations from a global conditioning vector and applies them uniformly in space. Our adapter (i) extracts factor-specific features through cross-attention rather than from a vector, and (ii) writes back \emph{spatially varying} updates, allowing different image regions (e.g., shadow boundaries vs.\ specular highlights) to be modulated differently for each factor.

\paragraph{Number of parameters.}
The adapter adds a small parameter overhead relative to the full model, while enabling factor-specific read--write modulation across the DPT-aligned feature hierarchy.

\Needspace*{5\baselineskip}
\subsection{Per-Factor Dense Heads}
\label{sec:supp_arch_heads}

Each factor uses a DPT-style dense head $\mathcal{H}_k$ that fuses tokens from four hierarchical levels, with adapter modulation applied at all four levels. All heads share the same DPT decoder hyperparameters and differ only in input tokens, output channels, and the activation on the final layer.

\paragraph{Albedo head.}
A non-negative linear-RGB output $\hat{\mathbf{A}}\!\in\!\mathbb{R}_{\geq 0}^{H\times W\times 3}$. Since albedo carries the most high-frequency texture, we additionally fuse multi-scale CNN features from a ResNeXt~\cite{xie2017aggregated} branch via gated addition at each DPT scale. The CNN branch operates on the per-view RGB and contributes pixel-aligned local cues; the gating is initialized to inject only $5\%$ of CNN features so that the multi-view tokens dominate at the start of training.
\rev{Concretely, the branch is a ResNeXt-101 ($32{\times}8$d) applied to each view's RGB, trained from scratch without ImageNet pretraining.
Its four feature scales (256/512/1024/2048 channels) are projected by $1{\times}1$ convolutions to 256/512/1024/1024 channels and fused into the corresponding DPT scales through the learned gates.
Each per-factor DPT head reads aggregator tokens from layers 4/11/17/23 and decodes them with a width-256 RefineNet-style decoder; the auxiliary depth and normal heads share this layout.}

\paragraph{Shading head.}
RGB shading $\hat{\mathbf{S}}\!\in\!\mathbb{R}_{\geq 0}^{H\times W\times 3}$ in HDR linear space. We use a softplus activation to keep predictions strictly positive while allowing values $>1$.

\paragraph{Residual head.}
$\hat{\mathbf{R}}\!\in\!\mathbb{R}^{H\times W\times 3}$ with no positivity constraint, since $\hat{\mathbf{R}}$ may locally compensate for small reconstruction errors as well as represent specular highlights and other non-diffuse effects.

\paragraph{Auxiliary geometry heads.}
Optional depth and surface-normal heads share the multi-view tokens without adapter modulation and follow VGGT's geometry head architecture.

\Needspace*{6\baselineskip}
\section{Training and Datasets}
\label{sec:supp_training_data}
\sectionaccent

\Needspace*{5\baselineskip}
\subsection{Training Details}
\label{sec:supp_training}

\paragraph{Optimization.}
AdamW~\cite{loshchilov2019decoupled} with peak LR $5\!\times\!10^{-5}$, weight decay $0.05$, cosine schedule with $5\%$ linear warm-up, and bfloat16 mixed precision. Gradient clipping at $1.0$. Trained on 4$\times$NVIDIA H100 (80\,GB).

\paragraph{Two-stage schedule.}
(i) \emph{Adapter warm-up} (200 epochs): backbone frozen; only adapters and dense heads are trained. This anchors factor-specific routing without disrupting the multi-view geometry prior. (ii) \emph{Joint fine-tuning} (300 epochs): backbone, adapters, and heads optimized jointly with the same loss weights.

\paragraph{Input setting.}
During training, for each multi-view sequence, we first randomly sample the number of input views $V$ from $[2,8]$, and then randomly select $V$ frames from the sequence when sufficient frames are available.
For the default quantitative setting in the main paper, MVID uses 5 input views with patch size 14.
A-CV is computed over the evaluation sequences: 5 views per Hypersim scene and 20 views per real-world scene, as described in Sec.~\ref{sec:supp_data}.
Each baseline is evaluated on the same test images and preprocessing protocol whenever applicable.

\paragraph{Resolution-aware fine-tuning.}
After the main training schedule, we fine-tune the final checkpoint at the evaluation resolution.
To fit the higher-resolution inputs within GPU memory, this stage uses fewer input views per batch while keeping the same losses and data mixture.
This reduces train--test resolution mismatch and improves high-frequency albedo and shading details, without changing the model architecture or evaluation protocol.

\paragraph{Loss weights.}
$\lambda_{\mathrm{intr}}\!=\!1.0$, $\lambda_{\mathrm{recon}}\!=\!0.5$, $\lambda_{\mathrm{cv}}\!=\!0.2$, $\lambda_{\mathrm{ada}}\!=\!0.05$ (with sub-weights $\lambda_{\mathrm{div}}\!=\!1.0,\,\lambda_{\mathrm{floor}}\!=\!0.1,\,\lambda_{\mathrm{l2}}\!=\!0.01$), $\lambda_{R}\!=\!0.05$, $\lambda_{\mathrm{aux}}\!=\!0.5$ ($\lambda_D\!=\!1.0,\,\lambda_N\!=\!0.5$). Table~\ref{tab:supp_cost} reports the resulting training and inference cost.

\begin{table}[t]
\centering
\caption{\textbf{Training and inference cost.}}
\label{tab:supp_cost}
\renewcommand{\arraystretch}{1.1}
\rowcolors{2}{tabrowgray}{white}
\resizebox{0.95\linewidth}{!}{
\begin{tabular}{l c c c c}
\toprule
\rowcolor{mvidlight}
& \textbf{Params} & \textbf{Train mem} & \textbf{Inference mem} & \textbf{Time (s/frame)}\\
\midrule
MVID & 2.2\,B & 71\,GB ($\times$4) & 12--18\,GB$^\dagger$ & 0.19 \\
\bottomrule
\end{tabular}}
\footnotesize{$^\dagger$Varies with enabled output heads and batch settings.}
\end{table}

\Needspace*{5\baselineskip}
\subsection{Loss Definitions}
\label{sec:supp_loss}

We give explicit forms for the loss terms summarized in the main paper. The main paper presents the compact objective; below we specify the implementation details used for scale handling, valid masks, and residual regularization.

\paragraph{Intrinsic supervision.}
For factors with ground truth, we supervise each component with a valid-pixel mask $\Omega$. Albedo is supervised in linear RGB with a detached per-channel least-squares scale $s_A\!\in\![0.7,1.3]$:
\begin{equation}
\mathcal{L}_A^{\mathrm{intr}}
=
\|s_A\hat{\mathbf{A}}-\mathbf{A}\|_{1,\Omega}
+
\lambda_{\nabla}
\|\nabla(s_A\hat{\mathbf{A}})-\nabla\mathbf{A}\|_{1,\Omega}.
\end{equation}
Diffuse shading is supervised in log-HDR space with a detached per-channel scale $s_S\!\in\![0.7,1.3]$:
\begin{equation}
\mathcal{L}_S^{\mathrm{intr}}
=
\|\log(1+s_S\hat{\mathbf{S}})-\log(1+\mathbf{S})\|_{1,\Omega}
+
\lambda_{\nabla}
\|\nabla\log(1+s_S\hat{\mathbf{S}})-\nabla\log(1+\mathbf{S})\|_{1,\Omega}.
\end{equation}
For residual supervision, we avoid applying an unsigned log transform directly to $\hat{\mathbf{R}}$, since residual values may be signed. We instead use a signed-log transform
\[
\phi(x)=\mathrm{sign}(x)\log(1+|x|)
\]
and define
\begin{equation}
\mathcal{L}_R^{\mathrm{intr}}
=
\|\phi(\hat{\mathbf{R}})-\phi(\mathbf{R})\|_{1,\Omega_R}
+
\lambda_{\nabla}
\|\nabla\phi(\hat{\mathbf{R}})-\nabla\phi(\mathbf{R})\|_{1,\Omega_R}.
\end{equation}
Here $\Omega_R$ excludes near-zero residual pixels from the gradient term to avoid emphasizing numerical noise in regions without non-diffuse effects. The total intrinsic loss is
\begin{equation}
\mathcal{L}_{\mathrm{intr}}
=
\mathcal{L}_A^{\mathrm{intr}}
+
\mathcal{L}_S^{\mathrm{intr}}
+
\mathcal{L}_R^{\mathrm{intr}} .
\end{equation}

\paragraph{Image formation reconstruction.}
Across all data sources, we enforce the residual image formation constraint in the same form as the main paper:
\begin{equation}
\mathcal{L}_{\mathrm{recon}}
=
\sum_i
\|\hat{\mathbf{A}}_i\odot\hat{\mathbf{S}}_i+\hat{\mathbf{R}}_i-\mathbf{I}_i\|_{1,\Omega}.
\end{equation}
For HDR synthetic buffers we optionally compute the same residual in log-compressed RGB by clamping the recomposed image to a small positive value before applying $\log(1+x)$; this affects only numerical conditioning and does not change the image formation model.

\paragraph{Cross-view consistency.}
For overlapping synthetic view pairs $(i,j)$ with reliable depth and camera poses, we define
\begin{equation}
\mathcal{L}_{\mathrm{cv}}
=
\sum_{(i,j)\in\mathcal{P}}
\|M_{ij}\odot(\hat{\mathbf{A}}_i-\mathcal{W}_{i\leftarrow j}(\hat{\mathbf{A}}_j))\|_1 .
\end{equation}
The overlap mask $M_{ij}$ requires: (i) the projected pixel lies inside the target image, (ii) reprojected depth agrees with target depth within a $5\%$ relative threshold, and (iii) both pixels are valid. During training, this loss is applied only on synthetic batches with reliable geometry, matching the main paper. For real-world cross-view evaluation, estimated geometry is used only to define valid reprojection overlaps.

\paragraph{Residual prior.}
On real-world batches without intrinsic ground truth, we discourage $\hat{\mathbf{R}}$ from absorbing diffuse appearance through a magnitude term and a low-frequency suppression term:
\begin{equation}
\mathcal{L}_R^{\mathrm{prior}}
=
\lambda_{\mathrm{mag}}\|\hat{\mathbf{R}}\|_1
+
\lambda_{\mathrm{lp}}\|\mathrm{LP}(\hat{\mathbf{R}})\|_1,
\end{equation}
where $\mathrm{LP}(\cdot)$ is a Gaussian-blurred copy of $\hat{\mathbf{R}}$ (kernel $\sigma\!=\!4$ pixels). This keeps the residual small and localized, consistent with its role as a non-diffuse correction term rather than a second appearance channel. We use $\lambda_{\mathrm{mag}}\!=\!0.01$ and $\lambda_{\mathrm{lp}}\!=\!0.05$.

\paragraph{Adapter regularization.}
Let $\bar{\mathbf{q}}_k^l$ denote the mean task token for factor $k$ at adapter layer $l$. We use
\begin{align}
\mathcal{L}_{\mathrm{div}} &=
\tfrac{1}{L_T'}\sum_l
\big\|
\mathrm{offdiag}\,
\mathrm{CosSim}(\bar{\mathbf{q}}^{l}_A,\bar{\mathbf{q}}^{l}_S,\bar{\mathbf{q}}^{l}_R)
\big\|_2^2,\\
\mathcal{L}_{\mathrm{floor}} &=
\max\!\big(0,\,
\tau_\alpha-\tfrac{1}{|K|L_T'}\sum_{k,l}\alpha^{l}_{k}
\big),\\
\mathcal{L}_{\mathrm{l2}} &=
\tfrac{1}{|K|L_T'}\sum_{k,l}(\alpha^{l}_{k})^2,
\end{align}
with $\tau_\alpha\!=\!0.02$, $L_T'\!=\!4$ adapter layers and $K\!=\!\{A,S,R\}$. The adapter regularization is
\begin{equation}
\mathcal{L}_{\mathrm{ada}}
=
\lambda_{\mathrm{div}}\mathcal{L}_{\mathrm{div}}
+
\lambda_{\mathrm{floor}}\mathcal{L}_{\mathrm{floor}}
+
\lambda_{\mathrm{l2}}\mathcal{L}_{\mathrm{l2}}.
\end{equation}

\begin{table}[t]
\centering
\caption{\textbf{Anchors against degenerate factorizations.}
Each constraint addresses a different ambiguity of the residual image formation model.}
\label{tab:supp_degeneracy_anchors}
\renewcommand{\arraystretch}{1.08}
\resizebox{0.95\linewidth}{!}{
\begin{tabular}{l l}
\toprule
\textbf{Anchor} & \textbf{Degeneracy discouraged} \\
\midrule
Synthetic intrinsic supervision 
& Factor swapping and scale drift \\

Cross-view albedo consistency 
& Shadows, highlights, or exposure changes absorbed into albedo \\

Residual prior 
& Residual becoming a free image-reconstruction channel \\

Factor-specific decoding 
& Leakage among material, illumination, and non-diffuse cues \\
\bottomrule
\end{tabular}}
\end{table}

\paragraph{Anchoring against trivial factorizations.}
The reconstruction equation $\mathbf{I}=\hat{\mathbf{A}}\odot\hat{\mathbf{S}}+\hat{\mathbf{R}}$ alone admits degenerate solutions, e.g., $\hat{\mathbf{A}}=\mathbf{I},\hat{\mathbf{S}}=\mathbf{1},\hat{\mathbf{R}}=\mathbf{0}$ or $\hat{\mathbf{A}}=\mathbf{1},\hat{\mathbf{S}}=\mathbf{I},\hat{\mathbf{R}}=\mathbf{0}$.
MVID avoids these degeneracies through complementary anchors, summarized in Table~\ref{tab:supp_degeneracy_anchors}.
First, synthetic intrinsic supervision fixes the semantic roles and scales of the three factors, preventing the network from arbitrarily swapping albedo, shading, and residual explanations.
Second, cross-view albedo consistency penalizes view-dependent shadows, highlights, and exposure changes when they are absorbed into $\hat{\mathbf{A}}$, since the same surface point should have stable reflectance across views.
Third, the residual prior discourages $\hat{\mathbf{R}}$ from becoming a free image-reconstruction channel by keeping it sparse, small in magnitude, and low-frequency suppressed.
Finally, factor-specific decoding further separates the cues used by each branch, reducing leakage among material color, illumination, and non-diffuse effects.
On unlabeled real-world batches, the reconstruction loss therefore adapts the shared model to real appearance statistics, while these anchors preserve a meaningful decomposition.

\Needspace*{5\baselineskip}
\subsection{Datasets and Splits}
\label{sec:supp_data}

Table~\ref{tab:supp_data_full} summarizes the datasets, splits, supervision types, and usage in training and evaluation.

\begin{table*}[t]
\centering
\caption{\textbf{Dataset usage and splits.}
We use synthetic datasets for intrinsic supervision and real-world sequences for image-formation supervision.
``Full A/S/R'' denotes albedo, RGB diffuse shading, and residual supervision;
``A-only'' denotes albedo supervision;
``Rec.'' denotes RGB reconstruction through the residual image formation constraint.
UAVLight is used mainly to broaden outdoor drone-view coverage rather than as the primary source of quantitative gains.}
\label{tab:supp_data_full}
\renewcommand{\arraystretch}{1.1}
\rowcolors{2}{tabrowgray}{white}
\resizebox{0.98\linewidth}{!}{
\begin{tabular}{l l c c c l}
\toprule
\rowcolor{mvidlight}
\textbf{Dataset} & \textbf{Domain} & \textbf{Train} & \textbf{Test} & \textbf{Supervision} & \textbf{Used for}\\
\midrule
Hypersim~\cite{Roberts2021Hypersim} 
& syn.\ indoor & 338 & 15 & Full A/S/R + Rec. + geom. & training, primary eval\\

InteriorVerse~\cite{zhu2022learning} 
& syn.\ indoor & 4{,}080 & 50 & A-only + geom. & training, cross-domain albedo eval\\

Structured3D~\cite{zheng2020structured3d} 
& syn.\ indoor & 3{,}500 & --- & A-only + geom. & training\\

UAV-Synth (ours) 
& syn.\ outdoor & 50 & --- & Full A/S/R + Rec. + geom. & RGB Rec. training\\

UAVLight-Real~\cite{uavlight} 
& real outdoor & 70 & --- & RGB Rec. only & RGB Rec. training\\

ScanNet++~\cite{yeshwanth2023scannet++} 
& real indoor & 462 & --- & RGB Rec. + estimated geom. & RGB Rec. training\\

RealEstate10K~\cite{zhou2018stereo} 
& real in/outdoor & 66{,}000 & --- & RGB Rec. + camera & RGB Rec. training\\

\midrule

MAW~\cite{wu2023measured} 
& real indoor & --- & all & measured albedo & zero-shot eval\\

Mip-NeRF 360~\cite{barron2022mipnerf360} 
& real outdoor & --- & sel. & RGB & real A-CV, qualitative comparison\\

Tanks \& Temples~\cite{knapitsch2017tanks} 
& real outdoor & --- & sel. & RGB & real A-CV, qualitative comparison\\

Deep Blending~\cite{hedman2018deep} 
& real indoor & --- & sel. & RGB & real A-CV, qualitative comparison\\
\bottomrule
\end{tabular}}
\end{table*}

\paragraph{Hypersim evaluation split construction.}
We evaluate on a fixed 15-scene subset of the official Hypersim test split, using 5 views per scene for quantitative evaluation.
The subset excludes scenes with invalid or unstable released intrinsic buffers that prevent reliable metric computation.
The selection is fixed before running any method and is independent of prediction quality.
The remaining 15 scenes (75 evaluation images in total) are unseen during training across all data sources.
All baselines are evaluated on the same scenes and evaluation images.

For the main Hypersim cross-view consistency evaluation, we use the same 5-view setting on held-out scenes unseen during training, avoiding overlap with the training scenes used for $\mathcal{L}_{\mathrm{cv}}$. For the input-view ablation in the main paper, we additionally vary the number of input views with $V\!\in\!\{1,2,3,5,8\}$.

\paragraph{Real-world cross-view evaluation.}
For real-world A-CV, we evaluate on 6 held-out scenes from Deep Blending, Mip-NeRF~360, and Tanks\,\&\,Temples, using 20 views per scene.
Camera poses and depths are estimated with COLMAP~\cite{colmap} and are used only to define valid reprojection overlaps for evaluation.
All methods are evaluated with the same geometry and valid-overlap masks.

\paragraph{Outdoor drone-view data.}
UAVLight and our UAV-Synth extension are included to increase the diversity of outdoor, long-range, and drone-view scenes.
We do not treat UAVLight as the main source of our quantitative improvement: the pretrained multi-view backbone is primarily trained on common ground-level imagery, and its prior is less specialized for UAV viewpoints.
Accordingly, the benefit of UAVLight is more visible in qualitative outdoor generalization and failure reduction than in large aggregate metric gains.
We include UAVLight results to demonstrate that the proposed residual image-formation decomposition remains applicable beyond indoor benchmarks, not to claim that UAVLight alone drives the final performance.

\paragraph{MAW evaluation.}
We evaluate all methods zero-shot on MAW: no MAW data is used for training or fine-tuning.
For multi-view input, we group available frames from each scene into non-overlapping 5-view chunks.
Predicted albedo maps from all methods are mapped back to the original image coordinates before applying the official MAW masks, and all scores are computed with the official MAW evaluator without modification.

\paragraph{Per-baseline preprocessing.}
For Colorful Diffuse, RGB-X, DiffusionRenderer, and DNF-Intrinsic, we use the authors' official pretrained checkpoints and inference code with default settings, and convert outputs to linear RGB (sRGB only for visualization). For IDArb, MAIR, and MAIR++ we feed the same multi-view input. For methods predicting only $A,S$ (no $R$), we derive $R=I-A\odot S$ and mark these residual numbers with $\dagger$ because they are not directly comparable to methods that explicitly predict residuals.

\Needspace*{5\baselineskip}
\rev{\subsection{Training-Data Fairness}
\label{sec:supp_data_fairness}

MVID is not trained on the same data as prior methods, since the compared paradigms differ fundamentally (diffusion priors, object-centric renders, PBR factors), and an exactly data-matched comparison is impractical.
Three observations support the fairness of the comparison.
First, our labeled intrinsic supervision uses $\sim$0.4M synthetic images, an order of magnitude fewer than recent generative baselines: IDArb trains on 5.7M labeled renderings and Diffusion Renderer on $\geq$3.6M labeled frames plus auto-labeled real videos, while RGB-X and DNF-Intrinsic use less labeled data but inherit Stable-Diffusion pretraining.
Second, our additional data is unlabeled posed RGB, exploited only through image-formation self-supervision; this is a capability of the method rather than extra annotation, analogous to Diffusion Renderer's use of auto-labeled real videos.
Third, Table~4 of the main paper quantifies its effect: removing real-world RGB training barely changes synthetic accuracy (albedo PSNR 24.72 to 24.45) but costs real-world consistency (A-CV 25.14 to 22.70), i.e., our synthetic-benchmark advantage does not come from data scale.
Finally, the zero-shot MAW evaluation (Sec.~\ref{sec:supp_maw}), where no method trains on MAW, further supports the comparison under equal footing.}

\begin{figure*}[t]
\centering
\includegraphics[width=0.98\textwidth]{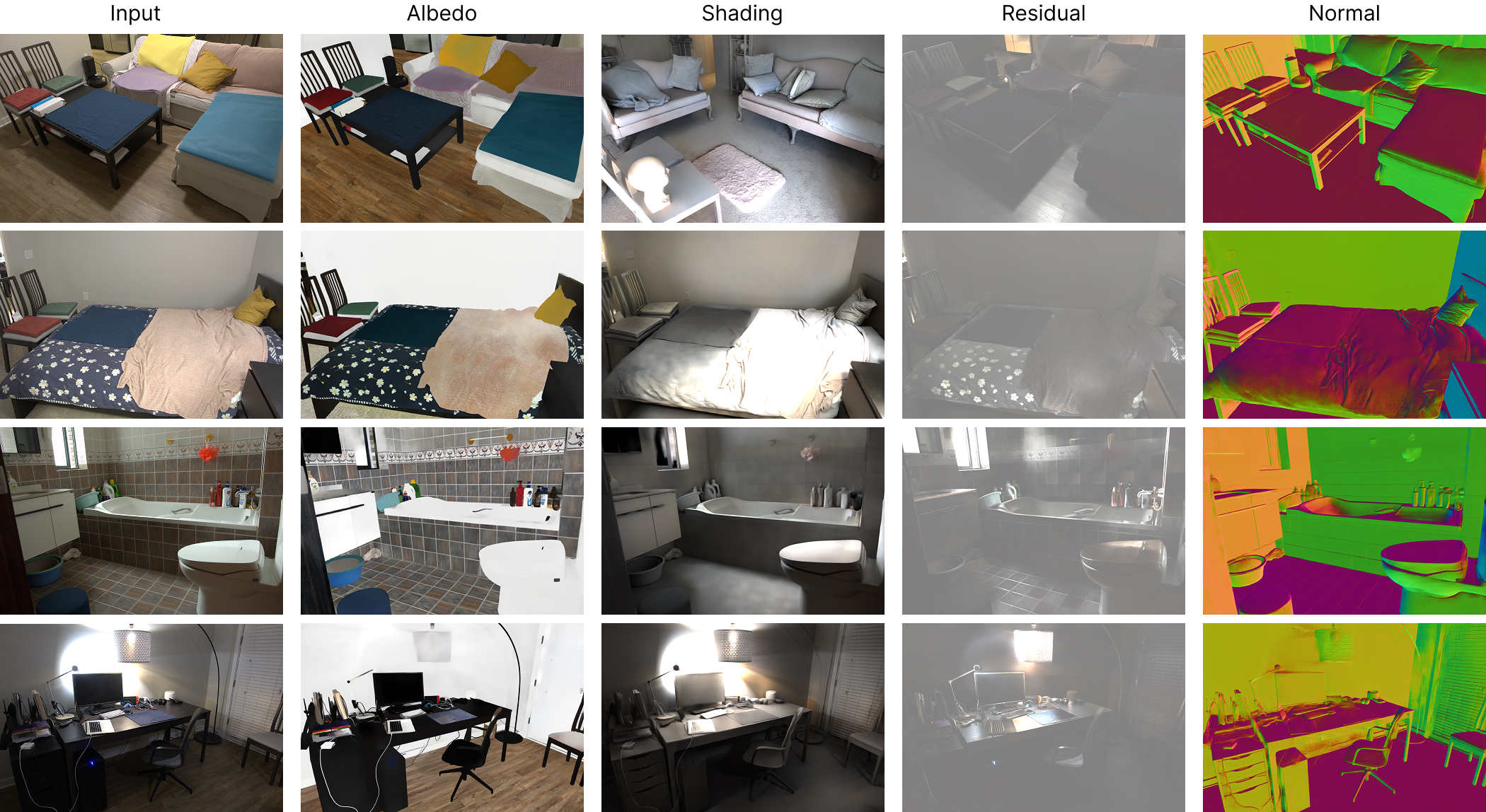}
\caption{\textbf{Additional decomposition results on MAW~\cite{wu2023measured}.}
MVID produces plausible real-world albedo and diffuse shading under measured-albedo evaluation scenes, with reduced shadow leakage in most regions.}
\label{fig:supp_maw_compare}
\vspace{-0.2cm}
\end{figure*}

\Needspace*{6\baselineskip}
\section{Evaluation Protocols}
\label{sec:supp_eval}
\sectionaccent

\Needspace*{5\baselineskip}
\subsection{Intrinsic Metrics}

\paragraph{Per-frame.}
On valid pixels we report PSNR (dB) and LPIPS (AlexNet) in linear RGB. Predictions are scale-aligned to ground truth using a single global per-channel least-squares scale within $[0.7,1.3]$; the same alignment is used for all baselines.

\paragraph{Cross-view consistency.}
Given two views $i,j$ and an overlap mask $M_{ij}$, we define
\begin{equation}
\mathrm{CV\text{-}PSNR}(\mathbf{F})
=
\mathrm{PSNR}\!\left(
M_{ij}\odot\hat{\mathbf{F}}_i,\;
M_{ij}\odot\mathcal{W}_{i\leftarrow j}(\hat{\mathbf{F}}_j)
\right),
\end{equation}
and average over all valid pairs in the test split. We report A-CV by setting $\mathbf{F}=A$. On Hypersim, the main evaluation uses 15 scenes with 5 views per scene and ground-truth depth and poses. On real-world data, the evaluation uses 6 held-out scenes with 20 views per scene and COLMAP-estimated poses and depth.
Hypersim A-CV uses the same 15 held-out scenes and 5-view setting as the main evaluation.

\Needspace*{5\baselineskip}
\subsection{Geometry Metrics}
Depth is evaluated with median-scale alignment using AbsRel and $\delta_1$. Normals are evaluated after rotation into the reference camera frame, using mean angular error and Acc@$11.25^{\circ}$.

\begin{figure*}[t]
\centering
\includegraphics[width=0.98\textwidth]{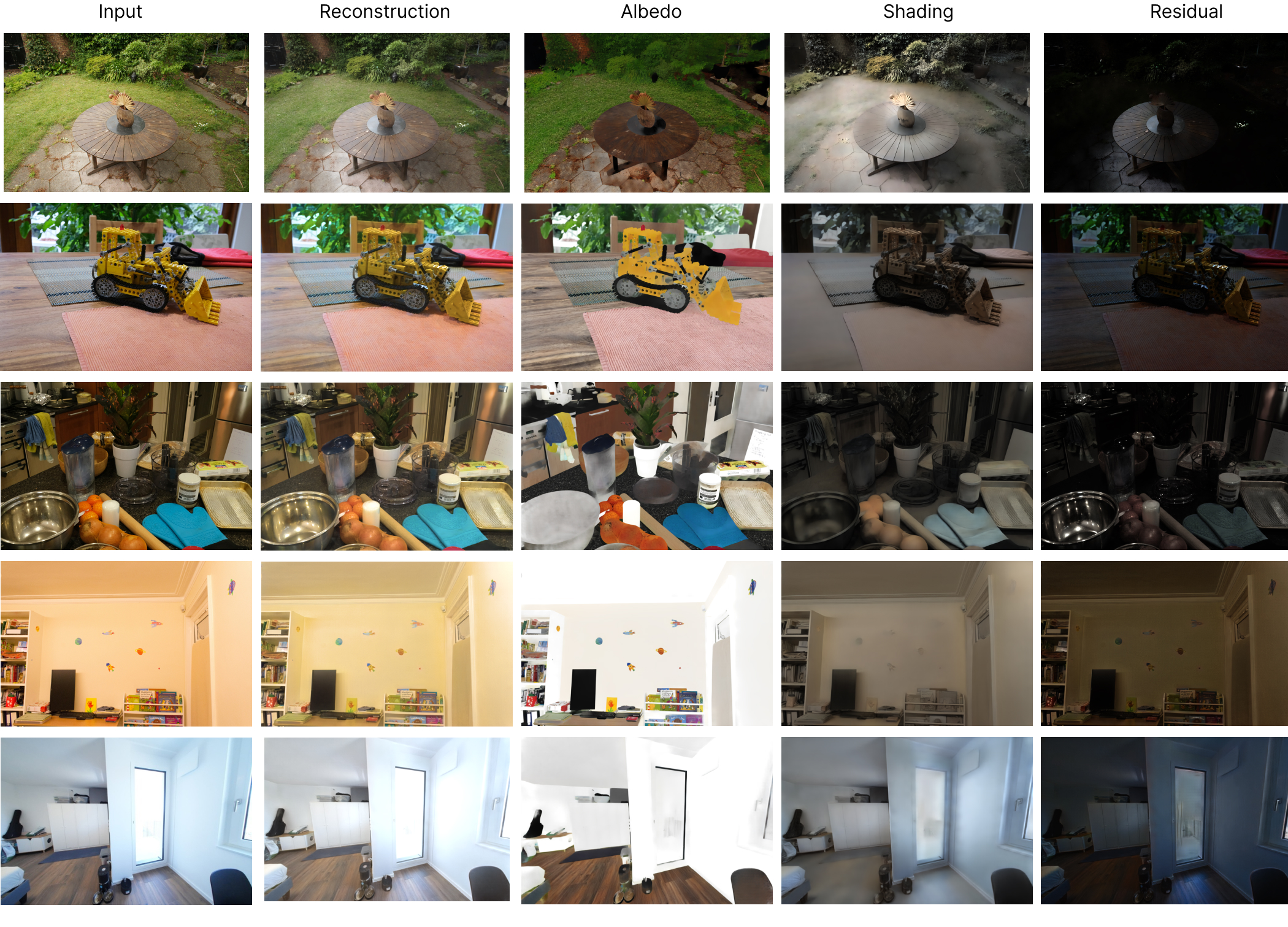}
\caption{\textbf{Additional real-world decomposition results.}
We show examples from Mip-NeRF~360~\cite{barron2022mipnerf360}, Deep Blending~\cite{hedman2018deep}, and ScanNet++~\cite{yeshwanth2023scannet++}.
MVID produces view-consistent image-domain factors on diverse real-world captures.}
\label{fig:supp_realworld_compare}
\end{figure*}

\Needspace*{6\baselineskip}
\section{Additional Quantitative Results}
\label{sec:supp_extra_quant}
\sectionaccent

\begin{figure*}[ht]
\centering
\includegraphics[width=0.95\textwidth]{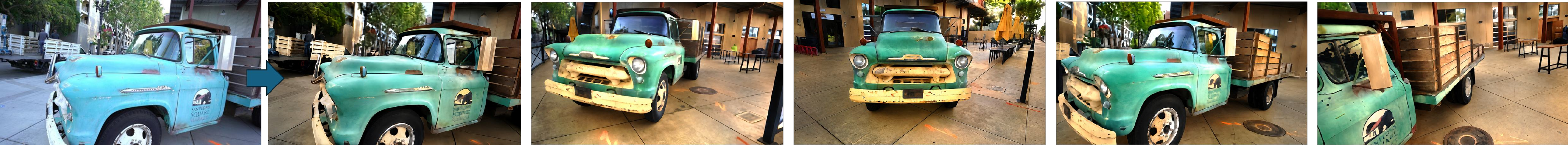}
\caption{
\textbf{Multi-view consistent appearance editing.}
MVID decomposes multi-view images into albedo, diffuse shading, and a non-diffuse residual with cross-view consistency.
By modifying only the diffuse shading appearance (e.g., intensity, color temperature, or shadow-like cues), we achieve coherent image-space edits across viewpoints while preserving albedo and residual components.
}
\label{fig:multiview_consistent_relight}
\end{figure*}

\begin{figure*}[t]
\centering
\includegraphics[width=0.99\textwidth]{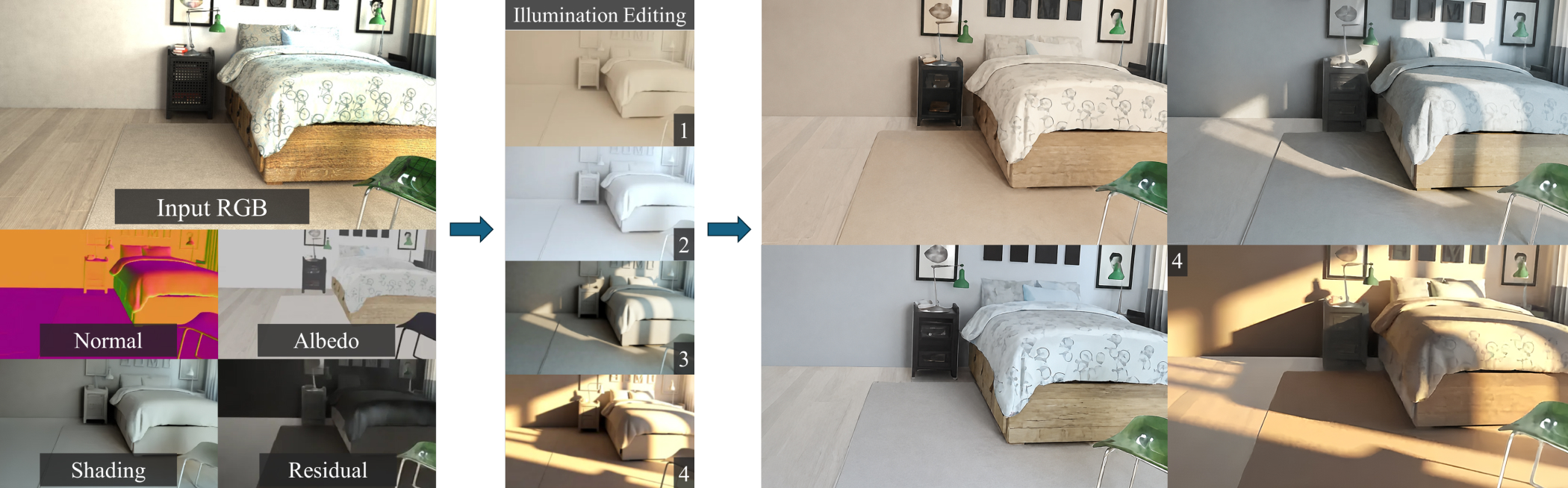}
\caption{
\textbf{Applications of MVID.}
MVID enables image formation decomposition for image-space illumination editing.
By editing the diffuse shading and residual while keeping albedo fixed, MVID produces plausible appearance edits with controllable color temperature and directional shading effects.
}
\label{fig:application1}
\end{figure*}

\Needspace*{5\baselineskip}
\subsection{Real-world Albedo Accuracy (MAW)}
\label{sec:supp_maw}

\input{table/tab_maw_eval}

On the MAW benchmark of measured real-world albedo, MVID achieves the strongest overall scores among the evaluated feed-forward baselines in our protocol (Table~\ref{tab:maw}), without training or fine-tuning on MAW.
\rev{The revision adds DNF-Intrinsic~\cite{zheng2025dnf} under the same official protocol (intensity 0.80, chromaticity 3.76, vs.\ MVID's 0.46 and 3.30).}
This supports the claim that the decomposition learned from synthetic supervision and weak real-world image-formation training transfers to measured real-world albedo.
When using published baseline numbers, we report them only when the evaluation protocol matches ours; otherwise we rerun the official checkpoints.
Fig.~\ref{fig:supp_maw_compare} further shows that MVID produces visually plausible albedo on real scenes, with reduced shadow leakage and more stable material colors compared with single-view baselines.

\begin{figure*}[t]
\centering
\includegraphics[width=0.98\textwidth]{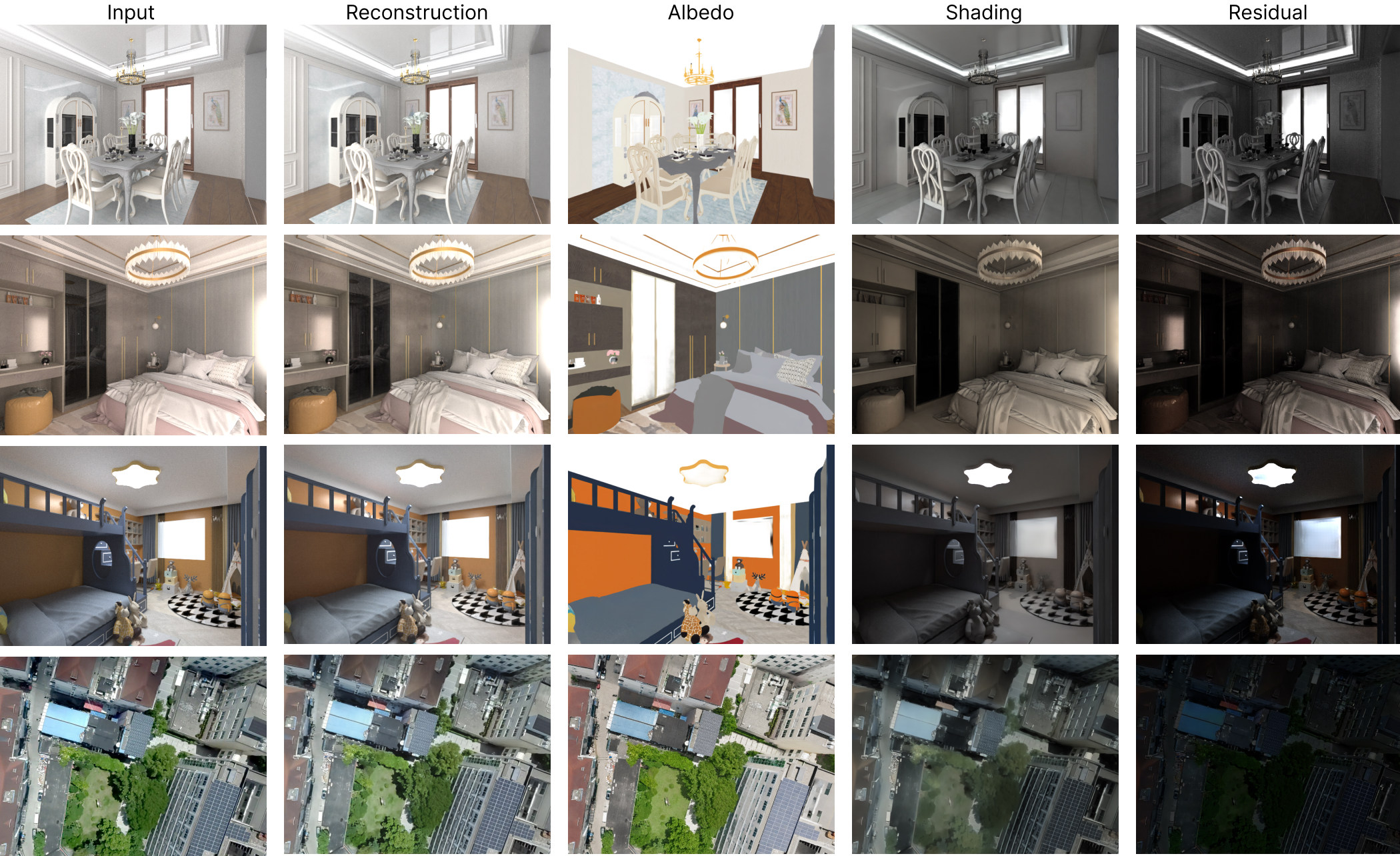}
\caption{\textbf{Additional decomposition results on InteriorVerse~\cite{zhu2022learning} and UAVLight~\cite{uavlight}.}
UAVLight provides outdoor drone-view scenes with larger depth ranges and stronger cast shadows.
These examples support qualitative generalization beyond indoor scenes, although UAVLight is not the primary source of our quantitative gains.}
\label{fig:supp_indoor_outdoor_compare}
\end{figure*}

\Needspace*{5\baselineskip}
\subsection{Backbone-only Sanity Check}
\label{sec:supp_backbone}
\begin{figure*}[h]
\centering
\includegraphics[width=0.98\textwidth]{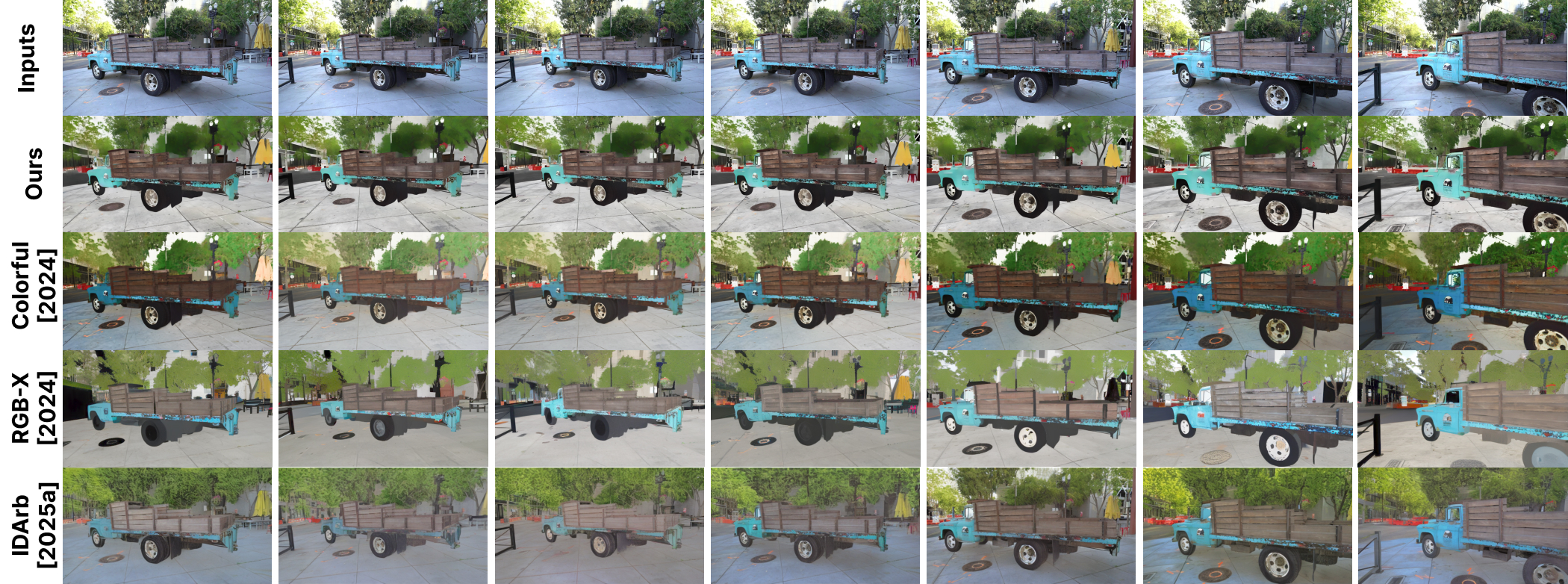}
\caption{
\textbf{Additional multi-view albedo stability on real-world captures.}
We show albedo predictions on a Truck sequence from Tanks\,\&\,Temples across multiple viewpoints.
Existing methods exhibit view-dependent albedo drift, inconsistent shadow removal, and unstable material appearance across frames.
In contrast, MVID produces more stable albedo while preserving scene structure and material appearance, further supporting its cross-view consistency on real-world captures.
}
\label{fig:MoreResults_albedo}
\end{figure*}

\begin{figure}[t]
\centering
\includegraphics[width=0.98\linewidth]{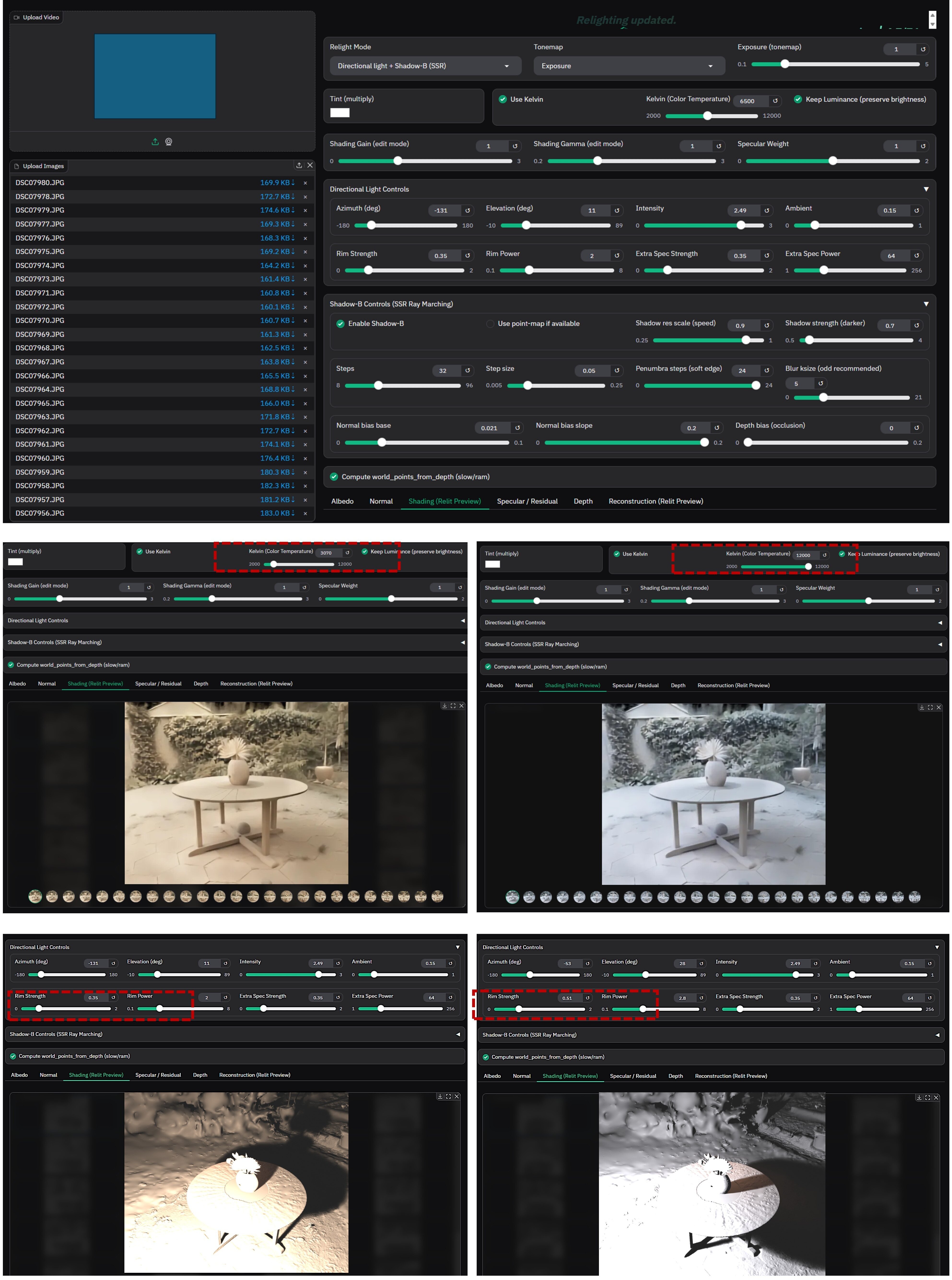}
\caption{
\textbf{Interactive image-space editing tool.}
Our Gradio-based interface exposes MVID's predicted albedo, diffuse shading, residual, depth, and normal buffers as editable layers.
Users can adjust shading intensity, color temperature, gamma, and directional shading cues, and apply the same edit consistently across views.
}
\label{fig:supp_tool_ui}
\vspace{-0.2cm}
\end{figure}

A natural concern is whether MVID's gains come solely from a stronger backbone.
We therefore train two variants: (B-1) replacing VGGT with a vanilla ViT-L initialized from DINOv2~\cite{oquab2024dinov2}, without multi-view pretraining, and (B-2) freezing the VGGT backbone throughout training.
On Hypersim, (B-1) reaches A PSNR $22.05$ / A-CV $24.30$, while (B-2) reaches $23.62$ / $25.74$, both below the full model.
These results indicate that MVID benefits from the pretrained multi-view representation, but its final gains also require adapting this representation with our factor-aware decoder and image-formation training objective.

\Needspace*{5\baselineskip}
\subsection{Cross-domain Albedo on InteriorVerse}
\label{sec:supp_interiorverse}

We further evaluate albedo on held-out InteriorVerse scenes that are disjoint from the InteriorVerse training scenes.
This setting tests cross-domain generalization from our mixed training data to a different synthetic indoor renderer.
A-PSNR is computed in linear RGB under the same alignment protocol as Hypersim.
As shown in Table~\ref{tab:supp_interiorverse_albedo}, MVID improves per-frame albedo quality over representative single-view and multi-view baselines.

\rev{%
\Needspace*{5\baselineskip}
\subsection{Temporal Stability Along Dense Trajectories}
\label{sec:supp_utv}

A-CV evaluates same-surface agreement between sparse overlapping views after geometric warping, which factors out camera motion.
On dense, nearly-aligned trajectories, temporal flickering can instead be measured directly by the \emph{unwarped temporal variance} (UTV) of the predicted albedo: the mean per-pixel variance computed over sliding 5-frame windows, with each method's predictions scale-normalized by their sequence mean to remove the global scale ambiguity of intrinsic decomposition.
We compute UTV on the dense 279-frame MipNeRF360-Kitchen sequence, processing every frame without subsampling (MVID runs in 30-frame chunks).
Table~\ref{tab:supp_utv} reports the results: the per-frame generative baseline RGB-X flickers as much as the raw input video itself, whereas MVID's albedo varies about $3\times$ less than the input, supporting temporal anchoring of material identity by the multi-view aggregator.

\begin{table}[t]
\centering
\caption{\textbf{Unwarped temporal variance (UTV) of predicted albedo} on the 279-frame MipNeRF360-Kitchen sequence (per-pixel, 5-frame windows, scale-normalized; lower is more temporally stable). The input row reports the variance of the raw video as a reference for how much the observation itself changes.}
\label{tab:supp_utv}
\renewcommand{\arraystretch}{1.08}
\resizebox{0.7\linewidth}{!}{
\begin{tabular}{lc}
\toprule
\textbf{Method} & \textbf{UTV}$\downarrow$ \\
\midrule
Input RGB (reference) & 0.176 \\
RGB-X~\cite{rgbx2024} (per-frame) & 0.177 \\
\textbf{MVID (Ours)} & \textbf{0.056} \\
\bottomrule
\end{tabular}}
\end{table}

\Needspace*{5\baselineskip}
\subsection{Robustness to Widening Camera Baselines}
\label{sec:supp_span}

To probe representation drift under widening baselines, we fix one anchor view per scene on the Hypersim test split (15 scenes, 5 views) and widen the baseline by sampling the supporting views farther apart, with frame spans of 4, 8, 12, and 16; the average adjacent-view overlap decreases from ${\sim}64\%$ to ${\sim}49\%$.
Table~\ref{tab:supp_span} reports albedo PSNR under the standard scale-aligned protocol: across all spans, the anchor-view PSNR varies within 0.02\,dB and the all-view mean within 0.2\,dB, i.e., no observable drift or token decay while views still overlap.

\begin{table}[t]
\centering
\caption{\textbf{Albedo PSNR under widening camera baselines} on the Hypersim test split (fixed anchor view; scale-aligned protocol). Accuracy is essentially flat as the frame span grows from 4 to 16.}
\label{tab:supp_span}
\renewcommand{\arraystretch}{1.08}
\resizebox{0.95\linewidth}{!}{
\begin{tabular}{lcccc}
\toprule
\textbf{Frame span} & \textbf{4} & \textbf{8} & \textbf{12} & \textbf{16} \\
\midrule
Anchor-view PSNR$\uparrow$ & 28.96 & 28.97 & 28.96 & 28.97 \\
All-view mean PSNR$\uparrow$ & 29.27 & 29.37 & 29.46 & 29.27 \\
\bottomrule
\end{tabular}}
\end{table}

\Needspace*{5\baselineskip}
\subsection{Residual Regularizer Ablation}
\label{sec:supp_residual_reg}

The residual regularizer $\mathcal{L}_R$ (Sec.~\ref{sec:supp_loss}) prevents $\hat{\mathbf{R}}$ from degenerating into a free image-reconstruction channel.
We provide two forms of evidence.
First, Fig.~\ref{fig:supp_residual_ablation} compares the predicted residual of a variant trained without $\mathcal{L}_R$ and of the full model on a real-world sequence: without the regularizer, the residual absorbs diffuse appearance and acts as a free reconstruction channel.
Second, the same comparison shows that under the full model $\hat{\mathbf{R}}$ stays small, sparse, and localized to non-diffuse effects, directly supporting the assumptions behind the regularizer.

\begin{figure}[t]
\centering
\includegraphics[width=0.98\linewidth]{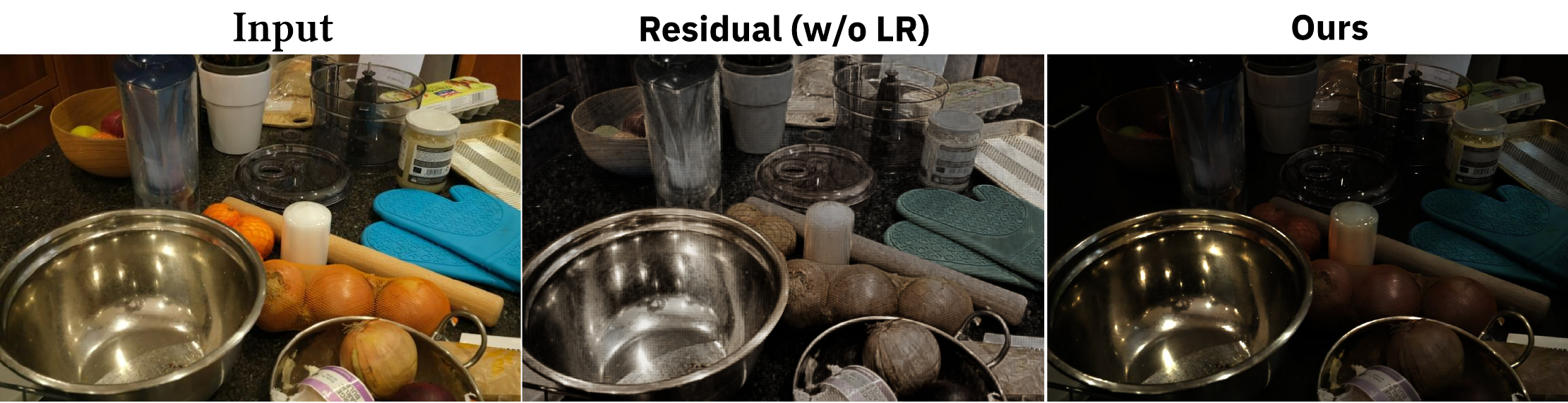}
\caption{\textbf{Visual ablation of the residual regularizer.}
Left to right: input, predicted residual without $\mathcal{L}_R$, and predicted residual of the full model.
Without $\mathcal{L}_R$, the residual absorbs large-scale diffuse appearance and acts as a free reconstruction channel; with $\mathcal{L}_R$, it stays small, sparse, and localized to non-diffuse effects such as the metallic highlights.}
\label{fig:supp_residual_ablation}
\end{figure}

\Needspace*{5\baselineskip}
\subsection{Local Refinement for Albedo: Ablation}
\label{sec:supp_local_refine}

We ablate the gated ResNeXt local-detail fusion in the albedo head (Sec.~\ref{sec:supp_arch_heads}).
Because the branch is trained jointly with the rest of the model, we retrain the model without it under the original protocol rather than disabling it at inference.
Fig.~\ref{fig:supp_local_refine} compares albedo predictions with and without the branch; quantitative results are reported in Table~4 of the main paper.

\begin{figure}[t]
\centering
\includegraphics[width=0.98\linewidth]{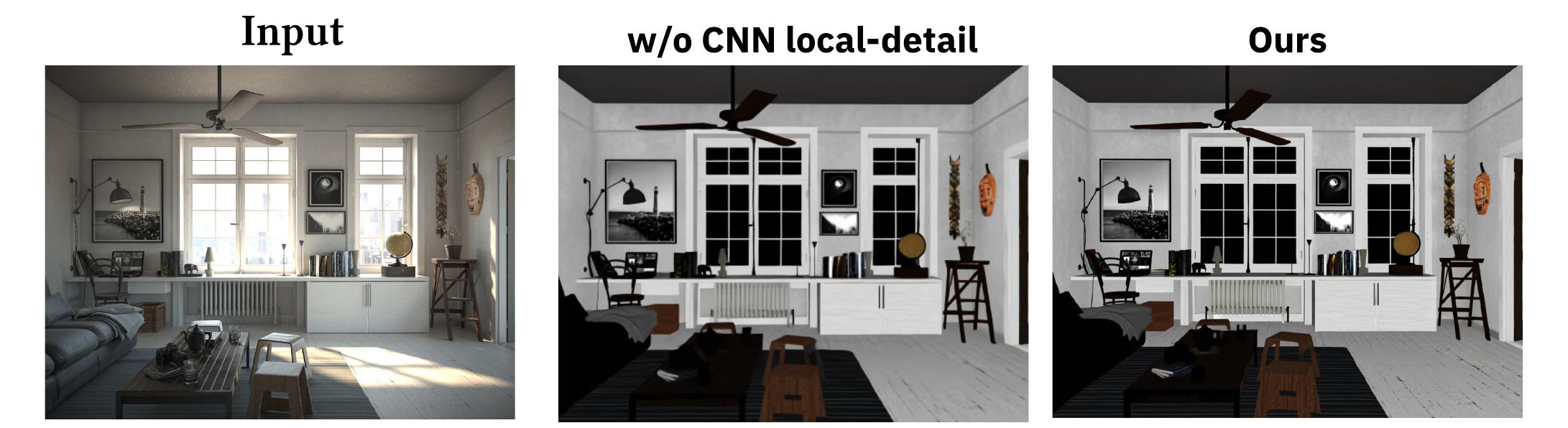}
\caption{\textbf{Ablation of the albedo local-refinement branch.}
Left to right: input, predicted albedo without the gated CNN local-detail fusion, and the full model.
Removing the branch softens high-frequency albedo texture, while cross-view stability is maintained by the multi-view tokens.}
\label{fig:supp_local_refine}
\end{figure}
}

\begin{table}[t]
\centering
\caption{\textbf{Cross-domain albedo on InteriorVerse.}
A-PSNR is computed on held-out InteriorVerse scenes in linear RGB.}
\label{tab:supp_interiorverse_albedo}
\renewcommand{\arraystretch}{1.08}
\resizebox{0.62\linewidth}{!}{
\begin{tabular}{llc}
\toprule
\rowcolor{mvidlight}
\textbf{Category} & \textbf{Method} & \textbf{A-PSNR}$\uparrow$ \\
\midrule
\rowcolor{tabrowgray}
Single-view & Colorful~\cite{careaga2024colorful} & 20.18 \\
Single-view & RGB-X~\cite{rgbx2024} & 21.04 \\
\rowcolor{tabrowgray}
Single-view & DiffR~\cite{DiffusionRenderer} & 21.67 \\
\midrule
Multi-view & IDArb~\cite{li2024idarb} & 19.42 \\
\rowcolor{tabrowgray}
Multi-view & \textbf{MVID (Ours)} & \textbf{22.18} \\
\bottomrule
\end{tabular}}
\vspace{-0.2cm}
\end{table}

\Needspace*{6\baselineskip}
\section{Adapter Visualization and Discussion}
\label{sec:supp_adapter_vis}
\sectionaccent

We complement the adapter visualization in Fig.~10(a) of the main paper with the following observations:

\begin{takeaway}
\small
\begin{enumerate}[leftmargin=1.5em, itemsep=2pt, topsep=0pt]
  \item[\textbf{(1)}] The three factors show distinct spatial response patterns despite sharing the same underlying multi-view tokens, supporting factor-specific cue preferences.
  \item[\textbf{(2)}] The responses are consistent across adapter layers and become more spatially refined at later layers.
  \item[\textbf{(3)}] Removing $\mathcal{L}_{\mathrm{div}}$ produces noisier but qualitatively similar responses, suggesting that diversity regularization stabilizes rather than solely enables the specialization.
\end{enumerate}
\end{takeaway}

\Needspace*{6\baselineskip}
\section{Discussion of Concurrent Work: MVInverse}
\label{sec:supp_mvinverse}
\sectionaccent

MVInverse~\cite{mvinverse2026} is concurrent with our work and also studies feed-forward multi-view appearance decomposition from sparse posed RGB inputs.
The two works are related at a high level but differ in output space, image formation model, and supervision.
Table~\ref{tab:supp_mvinverse_comparison} provides a structural comparison to clarify the relationship, rather than a direct numerical comparison.

\begin{table*}[t]
\centering
\caption{\textbf{MVID vs.\ concurrent MVInverse:} a structural comparison based on the public preprint.}
\label{tab:supp_mvinverse_comparison}
\renewcommand{\arraystretch}{1.25}
\setlength{\tabcolsep}{8pt}
\resizebox{0.98\linewidth}{!}{
\begin{tabular}{p{2.5cm} p{6.8cm} p{6.8cm}}
\toprule
\rowcolor{mvidlight}
\textbf{Aspect} & \textbf{MVInverse}~\cite{mvinverse2026} & \textbf{MVID (Ours)} \\
\midrule
\textbf{Goal} & Multi-view inverse rendering with BRDF-style material outputs & Multi-view intrinsic decomposition with image-domain factors \\
\midrule
\rowcolor{tabrowgray}
\textbf{Image formation} & Physically parameterized material and lighting model & Residual image formation $\mathbf{I}=\mathbf{A}\odot\mathbf{S}+\mathbf{R}$~\cite{careaga2024colorful} \\
\midrule
\textbf{Architecture} & Transformer-based multi-view architecture described in their preprint & VGGT-style multi-view aggregator + multi-layer factor-query adapter \\
\midrule
\rowcolor{tabrowgray}
\textbf{Supervision} & Synthetic supervision for material/BRDF-related quantities & Joint synthetic intrinsic supervision + unannotated real-world RGB reconstruction \\
\midrule
\textbf{Editing path} & Re-rendering through recovered parameters & Direct image-domain editing of predicted $A,S,R$ factors \\
\midrule
\rowcolor{tabrowgray}
\textbf{Evaluation focus} & Material/BRDF and geometry-related metrics & Per-factor PSNR/LPIPS, A-CV consistency, and reconstruction \\
\bottomrule
\end{tabular}}
\end{table*}

\paragraph{Why we do not include a direct head-to-head comparison.}
MVInverse is concurrent with our submission and, at the time of writing, remains an arXiv preprint rather than a finalized conference proceeding.
We therefore include it as concurrent work and provide a structural comparison in Table~\ref{tab:supp_mvinverse_comparison}.
A direct metric-level comparison would require locking down the final version, checkpoints, preprocessing, and evaluation protocol.
Moreover, MVInverse and MVID target different output spaces: BRDF-style material properties versus image-domain $A,S,R$ factors, so converting one output space into the other would introduce additional assumptions.

\paragraph{Summary.}
MVInverse and MVID share the broad motivation that feed-forward multi-view models can decompose appearance without per-scene optimization.
MVID is complementary: it focuses on view-consistent image-domain factorization under the residual image formation model, with direct support for image-space editing on real-world captures.

\Needspace*{6\baselineskip}
\section{Additional Qualitative Results}
\label{sec:supp_qual}
\sectionaccent

Fig.~\ref{fig:supp_maw_compare} shows additional real-world MAW decompositions.
Fig.~\ref{fig:supp_indoor_outdoor_compare} shows additional results on InteriorVerse and UAVLight, covering both synthetic indoor and outdoor drone-view scenes.
Fig.~\ref{fig:supp_realworld_compare} shows qualitative results on real-world multi-view captures from Mip-NeRF~360, Deep Blending, and ScanNet++.
Fig.~\ref{fig:MoreResults_albedo} further compares multi-view albedo stability on the Tanks\,\&\,Temples Truck sequence, showing that MVID reduces view-dependent albedo drift across real-world viewpoints.

\begin{figure}[t]
\centering
\includegraphics[width=0.98\linewidth]{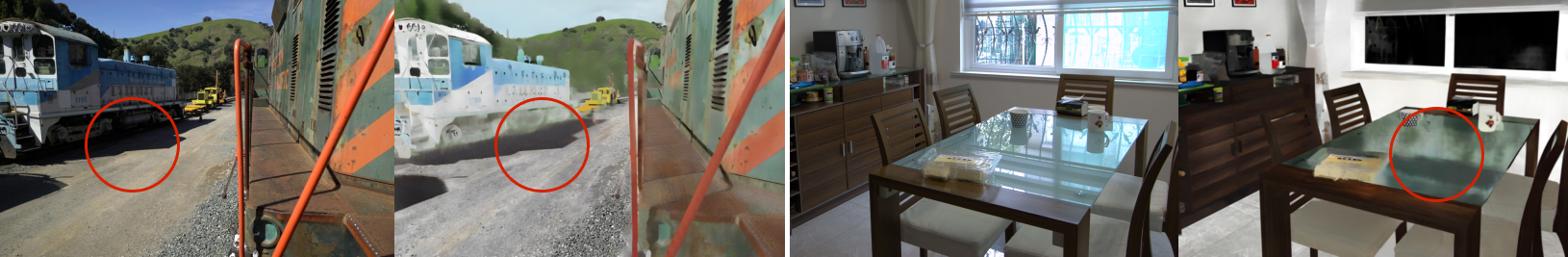}
\caption{
\textbf{Failure cases and limitations.}
Left: in extremely dark or near-saturated shadow regions, the observations contain very limited visual information, making decomposition inherently ambiguous without additional priors; this can occasionally lead to imperfect albedo recovery.
Right: MVID may also struggle with transparent glass and strong reflections, where transmission and view-dependent appearance cannot be fully modeled by the image-domain residual branch.
}
\label{fig:supp_failure}
\vspace{-0.6cm}
\end{figure}

\Needspace*{6\baselineskip}
\section{Application Details}
\label{sec:supp_apps}
\sectionaccent

Fig.~\ref{fig:supp_tool_ui} shows the interface of our interactive editing tool.
We implement an interactive editing tool that operates on MVID's predicted intrinsic factors and exposes them as editable layers to the user. The tool is built with a Gradio frontend and a CUDA inference backend.

\paragraph{Input modes.}
The tool accepts either posed multi-view images or a video. For video inputs, the tool samples frames to obtain a sparse view set, runs MVID, and uses the auxiliary camera/depth heads to estimate geometry when poses are not provided. The user can also override the sampled views by uploading individual frames.

\paragraph{Decomposition outputs.}
After a feed-forward pass, the tool exports aligned per-view buffers: albedo $\hat{\mathbf{A}}$, diffuse shading $\hat{\mathbf{S}}$, residual $\hat{\mathbf{R}}$, surface normals $\hat{\mathbf{N}}$, depth $\hat{\mathbf{D}}$, and the recomposed image $\hat{\mathbf{A}}\odot\hat{\mathbf{S}}+\hat{\mathbf{R}}$. These buffers are saved per frame so that the same edit can be applied across views.

\paragraph{Image-space illumination editing. \ref{fig:multiview_consistent_relight}}
Given a user-specified transformation $\mathcal{T}$ on $\hat{\mathbf{S}}$, the edited image of view $i$ is recomposed as
\begin{equation}
\widetilde{\mathbf{I}}_i
=
\hat{\mathbf{A}}_i \odot \mathcal{T}(\hat{\mathbf{S}}_i)
+
\hat{\mathbf{R}}_i.
\end{equation}
The tool exposes four operations:
\begin{itemize}[leftmargin=1.3em, itemsep=2pt, topsep=2pt]
\item \emph{Intensity scaling}: $\mathcal{T}(\hat{\mathbf{S}})=c\hat{\mathbf{S}}$, $c\!\in\![0.2,3.0]$.
\item \emph{Color temperature}: per-channel scaling $\mathcal{T}(\hat{\mathbf{S}})=\mathrm{diag}(r,g,b)\hat{\mathbf{S}}$ parameterized by a Kelvin slider.
\item \emph{Gamma}: $\mathcal{T}(\hat{\mathbf{S}})=\hat{\mathbf{S}}^\gamma$, $\gamma\!\in\![0.5,2.5]$.
\item \emph{Directional shading cues}: a Lambertian layer $L=\max(\hat{\mathbf{N}}\!\cdot\!\boldsymbol{\ell},0)\mathbf{c}_\ell$ is synthesized from predicted normals and a user-controlled direction/color, then mixed with $\hat{\mathbf{S}}$ as $\mathcal{T}(\hat{\mathbf{S}})=(1-\beta)\hat{\mathbf{S}}+\beta L$.
\end{itemize}

\paragraph{Screen-space geometry cues.}
For geometry-aware edits, we lift the predicted depth $\hat{\mathbf{D}}_i$ and normal $\hat{\mathbf{N}}_i$ into a per-view screen-space proxy.
User-specified directional cues are evaluated against this proxy to modulate $\hat{\mathbf{S}}_i$ before recomposition.
This remains an image-space editing operation rather than full physical relighting: no mesh, global illumination, or explicit BRDF optimization is used.
The proxy helps edits follow local surface orientation and major occlusion boundaries while preserving the decomposition-based editing workflow.

\paragraph{Cross-view consistency of edits.}
Because MVID produces stable albedo and applies the same transformation parameters to each view, edited results tend to remain coherent across nearby viewpoints.
This consistency depends on the quality of the predicted factors and geometry; we therefore present these results as image-space illumination editing rather than physically based relighting.

\paragraph{Specularity removal.}
The diffuse-only image $\hat{\mathbf{A}}\odot\hat{\mathbf{S}}$ provides an approximate specularity-reduced version of the input, following~\cite{careaga2024colorful}.

\Needspace*{6\baselineskip}
\section{Failure Cases and Limitations}
\label{sec:supp_failures}
\sectionaccent

\paragraph{Limited cues in extremely dark regions.}
As shown in Fig.~\ref{fig:supp_failure}, MVID can occasionally produce imperfect decomposition results in extremely dark regions, especially near saturated shadows or areas close to pure black.
In such cases, the observations contain very limited visual information, making it inherently difficult to reliably separate albedo, shading, and residual components without additional semantic or material priors.
These artifacts are typically localized and mainly occur in severely under-exposed regions.

\paragraph{Transparent glass and strong reflections.}
As shown in Fig.~\ref{fig:supp_failure}, MVID may also struggle with transparent or highly reflective regions, such as glass surfaces, where the observed appearance combines transmission, reflection, and view-dependent effects.
Our residual branch can absorb part of these effects, but it is not a parameterized BRDF or transparency model, so complex glass reflections may remain imperfectly decomposed.

\paragraph{Static-scene assumption.}
MVID assumes a static scene with sufficient view overlap.
Dynamic objects, rolling-shutter artifacts, or strong inter-view illumination changes can be absorbed by the residual or reduce cross-view consistency.

%% file: table/tab_maw_eval.tex
\begin{table}[t]
\centering
\scriptsize
\caption{\textbf{Real-world albedo accuracy on MAW}~\cite{wu2023measured}.
We report official intensity and chromaticity errors; $^*$ denotes grayscale-shading methods with fixed chromaticity.}
\label{tab:maw}
\setlength{\tabcolsep}{4pt}
\renewcommand{\arraystretch}{0.88}
\begin{tabular}{lcc}
\toprule
Method & Intensity ($\times 100$)$\downarrow$ & Chromaticity$\downarrow$ \\
\midrule
IIW~\citeyearpar{bell2014intrinsic}                       & 3.11 & 6.61 \\
CGIntrinsics~\citeyearpar{li2018learning}$^*$             & 1.72 & 6.56$^*$ \\
Sengupta~et~al.~\citeyearpar{sengupta2019neural}          & 2.17 & 6.39 \\
MGNet~\citeyearpar{zhu2022learning}                       & 1.44 & 4.94 \\
IID~\citeyearpar{kocsis2023intrinsicdiffusion}            & 1.13 & 5.35 \\
IntrinsicAnything~\citeyearpar{chen2024intrinsicanything} & 0.98 & 4.12 \\
IDArb~\citeyearpar{li2024idarb}                           & 0.87 & 4.08 \\
RGB-X~\citeyearpar{rgbx2024}                              & 0.48 & 3.65 \\
Colorful~\citeyearpar{careaga2024colorful}                & 0.54 & 3.37 \\
DiffR~\citeyearpar{DiffusionRenderer}                     & 0.51 & 3.98 \\
\rev{DNF~\citeyearpar{zheng2025dnf}}                      & \rev{0.80} & \rev{3.76} \\
\midrule
\textbf{MVID (Ours)}                                      & \textbf{0.46} & \textbf{3.30} \\
\bottomrule
\end{tabular}
\end{table}

%% file: section/1_introduction.tex

\section{Introduction}
Separating stable surface appearance from illumination- and view-dependent effects is a long-standing goal in intrinsic image decomposition~\cite{barrow1978intrinsic,land1971retinex}, with applications in image editing, relighting, and scene understanding.
The problem is challenging because real-world images are jointly determined by surface color, illumination, visibility, cast shadows, and other non-diffuse effects, making the decomposition inherently ambiguous.

Classical diffuse image formation, $I=A\odot S_{\text{gray}}$, assigns all observed appearance to albedo and shading.
This is insufficient for real-world images, where colored illumination, inter-reflections, specularities, visible light sources, and other non-diffuse effects cannot be cleanly explained by albedo and shading alone.
A recent alternative is the residual image formation model~\cite{careaga2024colorful}, $I=A\odot S+R$, where $A$ captures stable albedo, $S$ captures RGB diffuse illumination and shadows, and $R$ absorbs remaining non-diffuse or view-dependent appearance.
This formulation directly explains RGB observations without requiring explicit BRDF, lighting, visibility, or geometry estimation.

\begin{figure}[!tb] \centering \includegraphics[width=0.99\linewidth]{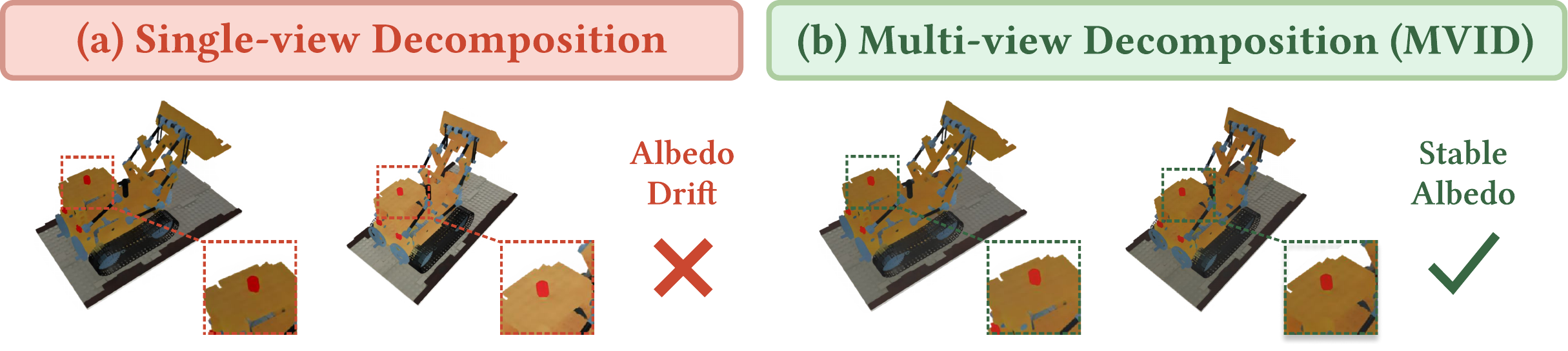}
\caption{
\textbf{Multi-view evidence reduces albedo drift.}
Applying MVID in a single-view setting (left) yields an albedo that drifts in color across views of the same scene, while the full multi-view MVID (right) produces a stable, view-consistent albedo (Test images are from TensoIR~\cite{jin2023tensoir}).
}
\label{fig:motivation}
\end{figure}

However, estimating these factors for multiple views is unstable.
Without scene-level context, a single-view method cannot reliably decide which appearance is material-stable and which is illumination-, visibility-, or view-dependent.
Applying single-view methods independently to consecutive views therefore causes albedo drift and inconsistent factor leakage across frames (Fig.~\ref{fig:motivation}).
Multi-view observations provide the missing context: the same surface point should keep consistent albedo, while \rev{diffuse shading explains illumination, shadows, and visibility changes, and the residual captures the remaining non-diffuse, view-dependent appearance}.

To this end, MVID first builds a scene-level context representation with a pretrained multi-view transformer, aggregating cross-view appearance and geometry cues to distinguish stable material appearance from illumination-, visibility-, and view-dependent variations.
Since albedo, shading, and residual rely on different cues, we further introduce a multi-layer factor query adapter.
The adapter decodes each factor from the shared representation with learned factor queries: albedo queries focus on material and color variations, while shading and residual queries respond more to shadows and specular highlights (Fig.~\ref{fig:ablation}).
Training such decomposition requires large amounts of data and benefits from real-world supervision.
The residual image formation model provides a natural self-supervised signal, as predicted factors can be directly recomposed by $\hat{I}=\hat{A}\odot\hat{S}+\hat{R}$ and compared with the input RGB.
We use this image-formation supervision on unlabeled real-world multi-view sequences, together with synthetic intrinsic supervision, cross-view albedo consistency, and a weak residual prior to prevent degenerate factorizations.

\rev{%
MVID maintains no explicit 3D representation at inference; its shared representation is instead geometry-initialized and geometry-supervised.
We verify that it behaves like geometry-grounded multi-view reasoning rather than per-frame filtering: a purely 2D backbone degrades accuracy and consistency, accuracy is stable under widening baselines, and gains grow with more input views (Sec.~\ref{sec:ablation}; backbone and baseline-widening analyses in the supplementary).%
}

Existing feed-forward multi-view methods pursue different output spaces.
IDArb~\cite{li2024idarb} is primarily evaluated for object-centric diffusion-based material prediction, while MAIR/MAIR++ \cite{choi2023mair,choi2025mairpp} recover inverse-rendering properties such as SVBRDF and lighting with MVS depth.
These parameter-space outputs are well-suited for physically based rendering, but \rev{self-supervised RGB reconstruction} is hindered by uncertain geometry, visibility, and lighting estimates.
In contrast, MVID targets \rev{view-stable} image-domain factors under the residual image formation model, enabling stable multi-view decomposition and direct \rev{self-supervised RGB reconstruction on unlabeled captures}.

We evaluate MVID on diverse synthetic and real-world datasets, including indoor benchmarks, Internet-scale multi-view videos, measured real-world albedo, and outdoor UAV captures.
These datasets cover Hypersim \cite{Roberts2021Hypersim}, InteriorVerse \cite{zhu2022learning}, MAW \cite{wu2023measured}, RealEstate10K \cite{zhou2018stereo}, ScanNet++~\cite{yeshwanth2023scannet++}, and UAVLight~\cite{uavlight}, with full details in Sec.~\ref{sec:exp_setup}.
We further augment the synthetic split of UAVLight with additional UAV scenes and intrinsic annotations, forming our UAV-Synth dataset for supervised training.
MVID outperforms single-view intrinsic, generative intrinsic, and multi-view inverse-rendering baselines in both per-frame decomposition quality and cross-view consistency, and supports practical applications such as \rev{image-space} illumination editing and specularity removal.


In summary, our contributions are:
\begin{itemize}[leftmargin=*, itemsep=2pt, topsep=2pt, parsep=0pt]
 
    \item
    We introduce \textbf{MVID}, the first feed-forward framework for multi-view intrinsic image decomposition under the residual image formation model, recovering \rev{view-consistent intrinsics}.

    \item
    We introduce a factor-specific read--write mechanism\rev{, in which learned per-factor queries \emph{read} factor-specific evidence from a shared multi-view representation and \emph{write} it back into the dense tokens, letting} albedo, shading, and residual decoders selectively extract information\rev{, reducing cross-factor leakage}.

    \item We develop a synthetic-real training strategy that combines intrinsic supervision with image-formation reconstruction on unannotated real-world sequences, improving in-the-wild decomposition and cross-view stability.

\end{itemize}

%% file: section/2_relatedwork.tex
\section{Related Work}

\textbf{Single-Image Intrinsic Image Decomposition.}
Intrinsic image decomposition traditionally explains an image as reflectance and shading~\cite{barrow1978intrinsic, land1971retinex}.
Classical methods rely on assumptions such as Lambertian reflectance, grayscale illumination, and local reflectance consistency~\cite{tappen2005intrinsic, bell2014intrinsic, chang2013non, Garces2022SurveyIntrinsic}.
Learning-based methods improve decomposition with data-driven priors and supervision~\cite{narihira2015direct, li2018learning, careaga2023ordinal}, but single-image prediction remains ambiguous since various combinations of reflectance, shading, geometry, illumination, and non-diffuse effects can explain the same observation. Recent work adopts residual image formation models that decompose images into albedo, colored diffuse shading, and non-diffuse residuals~\cite{lettry2016darn, careaga2024colorful}. By modeling these intrinsics, this formulation supports non-Lambertian decomposition and illumination-aware editing~\cite{careaga2025relighting}. Other methods use learned priors or generative diffusion models for perceptually plausible decompositions~\cite{rgbx2024, kocsis2023intrinsicdiffusion, luo2024intrinsicdiffusion, zheng2025dnf, chen2024intrinsicanything, sun2025ouroboros, kocsis2025intrinsix}, but their generated details may deviate from the input or fail to remain stable across views (Sec.~\ref{sec:sharpness_faithfulness}).
\rev{Beyond image-space priors, ReasonX guides intrinsic decomposition with multimodal language-model reasoning~\cite{reasonx}, and LumiNet combines latent intrinsics with diffusion models for indoor relighting~\cite{luminet2025}.}
Since these methods are inherently single-view, applying them independently to multiple views often yields unstable and inconsistent decompositions.

\begin{figure*}[ht] \centering \includegraphics[width=0.99\textwidth]{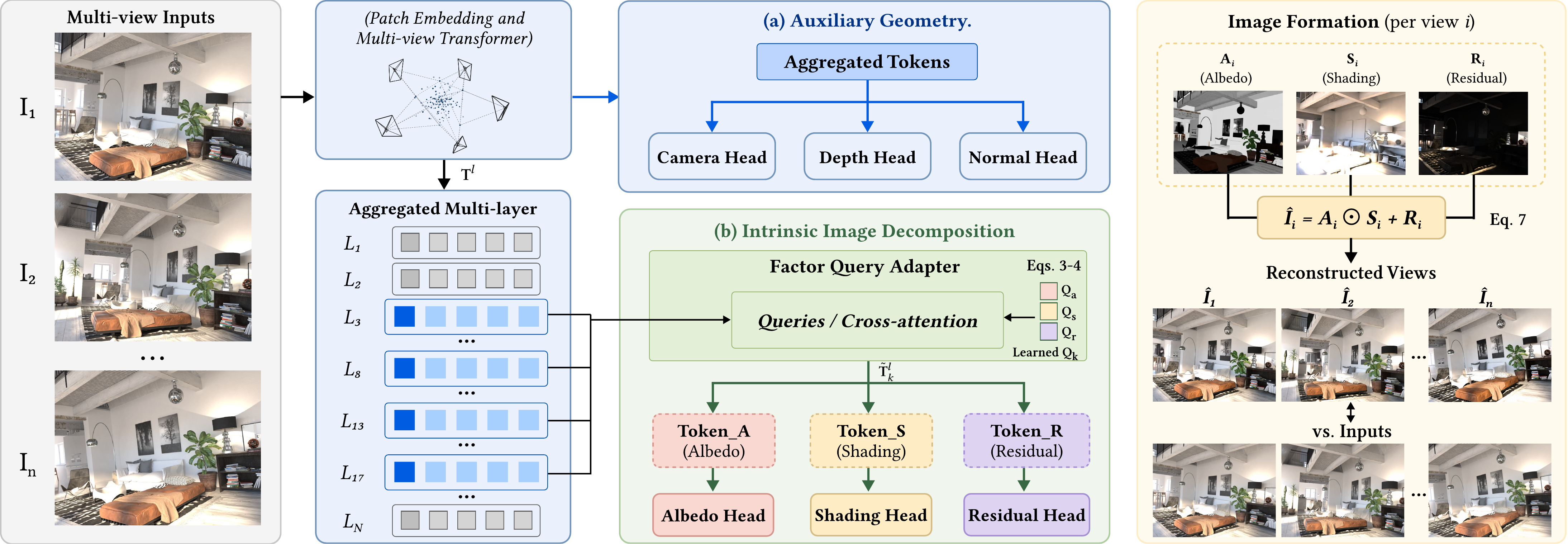}
\caption{
\textbf{MVID pipeline.}
Multi-view RGB inputs are aggregated by a transformer encoder into a hierarchy of cross-view tokens.
A dual-path decoder produces (a)~auxiliary geometry (camera, depth, normal) and (b)~intrinsic factors through a multi-layer factor query adapter that routes per-factor information into albedo, diffuse shading, and residual heads via learned queries.
The residual image formation model $\hat{I} = \hat{A} \odot \hat{S} + \hat{R}$ provides an image formation constraint, enabling joint training on synthetic and unlabeled real-world multi-view sequences.
}
\label{fig:pipeline}
\end{figure*}

\textbf{Multi-View Intrinsic Decomposition and Inverse Rendering.}
Multi-view observations provide additional constraints for decomposing appearance.
Early multi-view intrinsic methods exploit correspondences, geometry, or shadow cues to reduce single-image ambiguity, but typically rely on calibrated capture, fixed lighting assumptions, or per-scene optimization~\cite{duchene2015multiviewintrinsic}.
\rev{Complementary intrinsics from neural radiance fields and CNNs have also been combined for outdoor scene relighting, though still within a per-scene optimization framework~\cite{complementaryintrinsics}.}
Recent 3D representations enable scene-specific intrinsic decomposition by integrating intrinsic attributes into Gaussian Splatting and optimizing them from posed multi-view images~\cite{xing2024intrinsicgs, du2024gs, gao2024relightable3dgs, recap2025}, demonstrating the value of multi-view constraints but remaining tied to per-scene optimization.

Inverse rendering addresses a related but broader problem: recovering physical scene parameters such as geometry, materials, BRDF, and illumination~\cite{barron2014shape, karsch2011rendering}.
Optimization-based methods including NeRD~\cite{boss2021nerd}, NeRFactor~\cite{zhang2021nerfactor}, and TensoIR~\cite{jin2023tensoir} fit scene-specific neural or mesh-based representations from posed multi-view images.
Recent feed-forward methods avoid per-scene optimization: IDArb~\cite{li2024idarb} estimates intrinsic components from arbitrary numbers of views with diffusion priors; MAIR/MAIR++  \cite{choi2023mair, choi2025mairpp} estimate SVBRDF, normals, and 3D spatially-varying lighting from forward-facing multi-view inputs through a cascaded pipeline that requires MVS depth as an auxiliary input; \rev{MAGE~\cite{mage2025} lifts a single image to material-aware 3D by estimating multi-view G-buffers;} and concurrent to our work, MVInverse~\cite{mvinverse2026} predicts SVBRDF and normals from RGB sequences with alternating attention, supervised on synthetic data.
These methods primarily target physically parameterized scene recovery, typically estimating BRDF parameters.
In contrast, MVID targets image-domain intrinsic decomposition rather than full physical scene recovery.
It adopts the residual image formation model~\cite{careaga2024colorful} to predict view-consistent albedo, diffuse shading, and residual from sparse multi-view inputs, without explicit BRDF or lighting estimation.

\textbf{Multi-View Scene Representation.}
Recent feed-forward methods learn shared multi-view representations from sparse inputs for geometry estimation, correspondence reasoning, and novel-view synthesis~\cite{wang2024dust3r, cabon2025must3r, Wang2025VGGT}, showing that cross-view aggregation provides strong scene-level cues without per-scene optimization.
MVID builds on this paradigm: Our shared representation serves as the basis for view-consistent intrinsic factorization in the image domain, with factor-specific information routed through learned per-factor queries before dense decoding.

%% file: section/3_method2.tex
\section{Method}

\begin{figure*}[ht] \centering \includegraphics[width=0.99\textwidth]{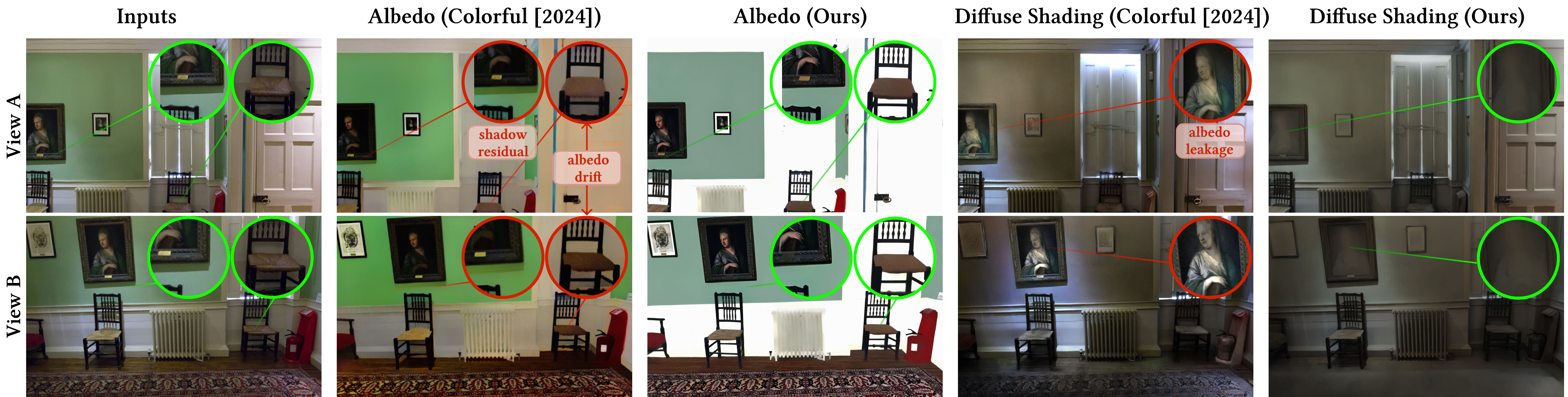}

\caption{
\textbf{Real-world decomposition comparison.} 
Colorful~\cite{careaga2024colorful} and MVID predict RGB diffuse shading under \(I=A\odot S+R\). When applied independently to nearby views, Colorful leaves shadow residuals and albedo drift, while its shading absorbs part of the material color.
MVID leverages multi-view evidence to obtain more stable albedo and cleaner diffuse shading that mainly captures illumination, shadows, and visibility-dependent effects.
}
\label{fig:realworld_decomp_compare}
\end{figure*}

Given sparse overlapping views of a static scene $\{I_i\}_{i=1}^{V}$, MVID predicts per-view albedo $A_i$, diffuse shading $S_i$, and residual $R_i$ under the residual image formation model
\begin{equation}
    I_i \approx A_i \odot S_i + R_i,
\end{equation}
where $\odot$ denotes element-wise multiplication. The pipeline (Fig.~\ref{fig:pipeline}) consists of three modules: a multi-view aggregator that produces a shared representation, multi-layer factor query adapters that route factor-specific information, and per-factor dense prediction heads.

\subsection{Shared Multi-View Representation}

\rev{Intuitively, rather than matching pixels across images explicitly, we fuse all views into one shared token representation from which each view's factors are decoded.}
We build on a transformer-based multi-view aggregator~\cite{Wang2025VGGT} to encode input views into \rev{this} shared multi-view representation. Each image is tokenized into patch embeddings, and alternating frame-wise and cross-view attention fuses information across overlapping views. This produces a hierarchy of dense token features
\begin{equation}
    \{\mathbf{T}^l\}_{l=1}^{L}, 
    \quad 
    \mathbf{T}^l \in \mathbb{R}^{V \times P \times C},
\end{equation}
where $L$, $V$, $P$, $C$ denote the numbers of aggregation layers, input views, tokens per view, and channel dimension, respectively. This representation is a hierarchy of cross-view aggregated tokens instead of an explicit 3D model. Since all per-view predictions are decoded from the same representation, they are implicitly coupled without requiring dense correspondence at inference time. 

\subsection{Multi-Layer Factor Query Adapter}
\label{sec:adapter}
Different intrinsic factors rely on different cues from the shared multi-view representation:
albedo benefits from view-stable material appearance, diffuse shading depends on illumination, geometry, and visibility, and non-diffuse residual captures localized view-dependent appearance.
Decoding all factors directly from the same shared tokens may cause factor leakage (Sec.~\ref{sec:ablation} and Table~\ref{tab:ablation}).

We address this with a \emph{multi-layer factor query adapter}.
\rev{Unlike a standard attention query, which is a projection of the input tokens, a \emph{factor query} is a learned per-factor probe: it \emph{reads} factor-specific evidence from the shared representation and \emph{writes} it back into the dense per-view tokens, giving each factor its own routing path against cross-factor leakage (Table~\ref{tab:ablation}).}
At adapter layer \(l\), we denote the shared patch tokens as
\(\mathbf{T}^l \in \mathbb{R}^{V \times P \times C}\).
For cross-view reading, we flatten the view and patch dimensions into
\(\bar{\mathbf{T}}^l \in \mathbb{R}^{(VP) \times C}\).
After factor-specific adaptation, the resulting tokens for factor
\(k\in\{A,S,R\}\) are denoted as
\(\tilde{\mathbf{T}}^l_k \in \mathbb{R}^{V \times P \times C}\).

\paragraph{Read.}
For each factor \(k\), a set of learned queries
\(\mathbf{Q}_k \in \mathbb{R}^{M \times C}\) cross-attends to the flattened multi-view memory:
\begin{equation}
    \mathbf{Z}^l_k = \mathrm{CrossAttn}\big(\mathbf{Q}_k,\, \bar{\mathbf{T}}^l\big),
\end{equation}
where we use \(M=8\) queries for each factor in all experiments.
Flattening across views allows each factor to gather evidence from the full multi-view context rather than from individual frames.

\paragraph{Write.}
The factor summary then writes back into the dense patch tokens through another cross-attention\rev{, in which the patch tokens \(\mathbf{T}^l\) act as the queries and the per-factor summaries \(\mathbf{Z}^l_k\) provide the keys and values}:
\begin{equation}
    \tilde{\mathbf{T}}^l_k =
    \mathbf{T}^l
    +
    \alpha^l_k
    \left(
    \mathrm{CrossAttn}(\mathbf{T}^l, \mathbf{Z}^l_k)
    -
    \mathbf{T}^l
    \right).
\end{equation}
Here \(\alpha^l_k\in\mathbb{R}\) is a learnable scalar gate shared across tokens and channels.
It is initialized to \(0.1\), so the adapter starts close to the identity mapping, following common practice in adapter modules~\cite{pfeiffer2020adapterhub}.
Special non-patch tokens \rev{(the aggregator's camera and register tokens)} are kept unchanged.

\paragraph{Hierarchical application.}
We apply the read--write operation at DPT-aligned aggregator layers
\(\mathcal{L}_{\mathrm{ada}}=\rev{\{4,11,17,23\}}\).
These layers match the feature scales consumed by the DPT-style dense heads.
The factor-conditioned token hierarchies \(\{\tilde{\mathbf{T}}^l_k\}_{l\in\mathcal{L}_{\mathrm{ada}}}\)
are then passed to the corresponding dense prediction head \(\mathcal{H}_k\).

\input{table/tab_dataset}

\paragraph{Relation to prior designs.}
Our adapter follows learned-query bottleneck designs such as Q-Former~\cite{li2023blip2}, DETR-style decoders~\cite{carion2020detr}, and Perceiver IO~\cite{jaegle2022perceiver}, but adapts read--write attention to multi-layer dense prediction and multi-factor decoding. Unlike FiLM-style conditioning~\cite{perez2018film}, which applies channel-wise affine modulation from a conditioning vector, our adapter uses cross-attention for both extraction and modulation. Without explicit attention supervision, the factor queries learn distinct visual preferences: albedo queries focus on reflectance cues, shading queries on shadow and illumination cues, and residual queries on specular highlights (Fig.~\ref{fig:ablation} (a)).

\input{table/tab_hypersim_intrinsic}

\begin{figure*}[t] \centering \includegraphics[width=0.99\textwidth]{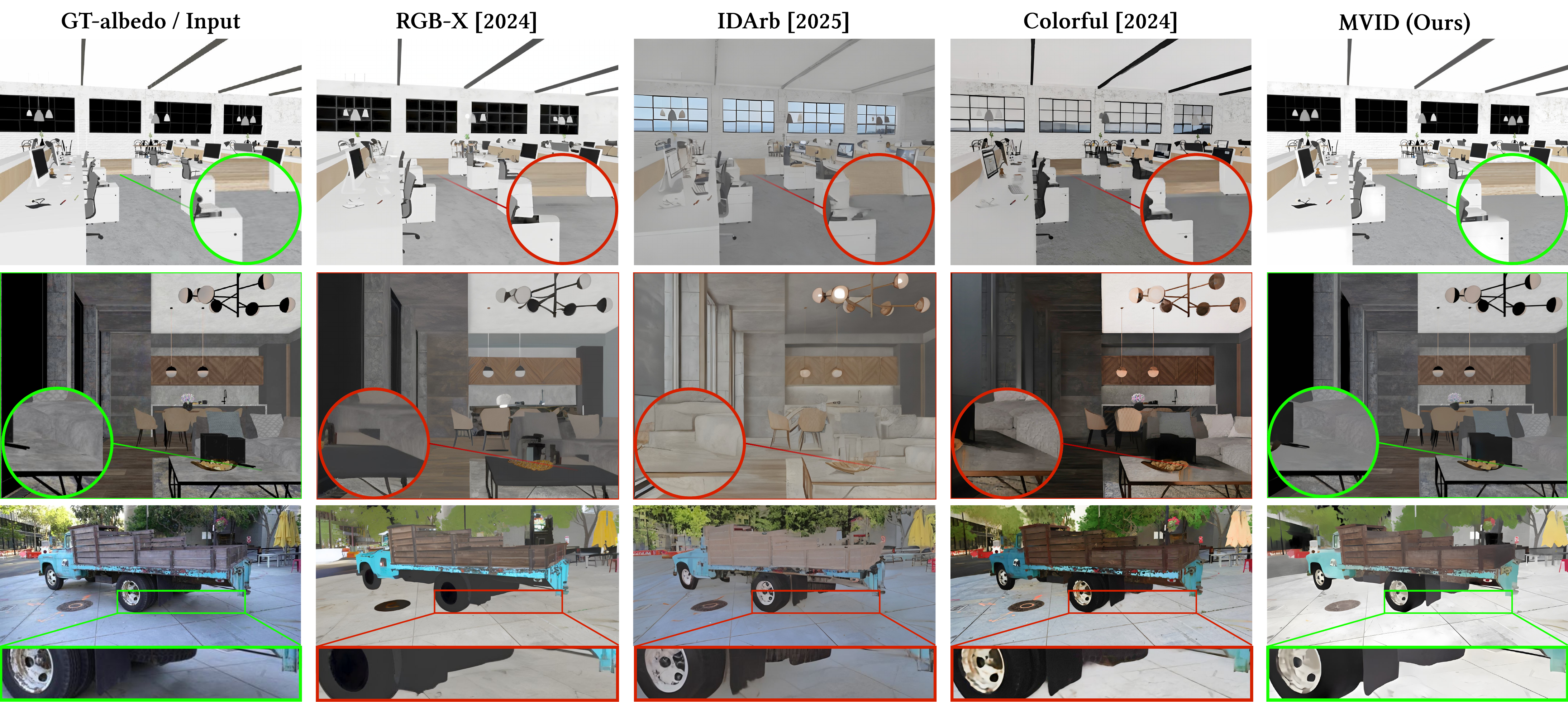}
\caption{
\textbf{Qualitative albedo comparison on Hypersim and MipNeRF 360.}
All results are shown in gamma space for visualization.
\rev{For the Hypersim scenes, the first column shows the ground-truth albedo.}
Compared with representative baselines, MVID produces cleaner albedo with less illumination leakage.
}
\label{fig:qual_albedo}
\end{figure*}

\subsection{Per-Factor Dense Prediction}

Given factor-conditioned token hierarchies, we predict intrinsics with per-factor DPT-style~\cite{ranftl2021vision} dense heads:
\begin{equation}
    \hat{A}_i = \mathcal{H}_A(\{\tilde{\mathbf{T}}^l_A\}),\;\;
    \hat{S}_i = \mathcal{H}_S(\{\tilde{\mathbf{T}}^l_S\}),\;\;
    \hat{R}_i = \mathcal{H}_R(\{\tilde{\mathbf{T}}^l_R\}).
\end{equation}
The shading and residual heads decode directly from their factor-conditioned tokens.

\paragraph{Local refinement for albedo.}
Albedo is the most texture-sensitive factor and benefits from pixel-aligned local evidence beyond patch-level multi-view tokens. We therefore fuse the albedo tokens with multi-scale CNN features~\cite{xie2017aggregated} through gated aggregation across the DPT hierarchy. The gates control the amount of local detail injected at each scale, preserving cross-view consistency while recovering high-frequency texture.

\paragraph{Auxiliary geometry.}
We optionally predict depth and surface normals from the shared representation as auxiliary outputs.
When supervision is available, these geometry signals provide structural regularization, particularly for diffuse shading.
Architectural details for all heads are provided in the supplementary.

\begin{figure*}[t]
\centering
\includegraphics[width=0.99\textwidth]{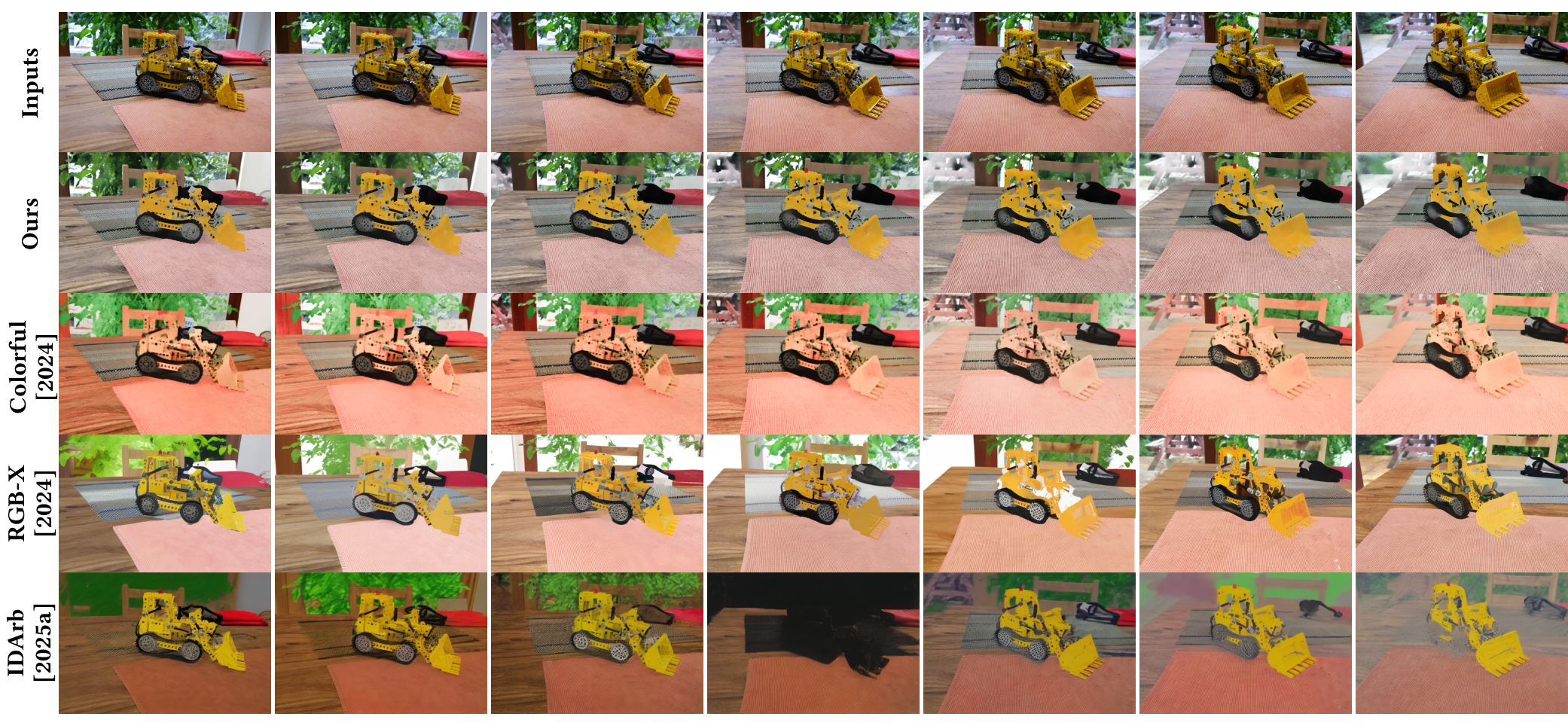}
\caption{
\textbf{Multi-view albedo stability on real-world captures.}
Albedo predictions on a Kitchen sequence sampled every other frame (shown in gamma space).
Single-view methods drift across views and diffusion-prior methods hallucinate plausible but incorrect details, while MVID remains stable and faithful.
}
\label{fig:real-world_captures}
\end{figure*}

\subsection{Training Objective}
\label{sec:training}

MVID is trained jointly on synthetic data with intrinsic annotations and real-world RGB sequences without intrinsic labels. The objective combines factor-level supervision when available, an image formation constraint, a cross-view consistency term, and lightweight regularizers for the adapter and the residual.

\paragraph{Factor-level supervision.}
On datasets with intrinsic ground truth, we directly supervise each predicted factor:
\begin{equation}
    \mathcal{L}_{\mathrm{intr}} = \sum_i \sum_{k \in \{A,S,R\}} 
    \big\| \hat{k}_i - k_i^{\mathrm{gt}} \big\|_1.
\end{equation}
This factor-level supervision anchors the decomposition and reduces degenerate factorizations.

\paragraph{Image formation constraint.}
Across all data sources, we enforce input consistency through the residual image formation model:
\begin{equation}
    \mathcal{L}_{\mathrm{recon}} = \sum_i \big\| \hat{A}_i \odot \hat{S}_i + \hat{R}_i - I_i \big\|_1.
\end{equation}
On real-world RGB sequences without intrinsic annotations, this constraint provides the primary supervision signal, allowing MVID to leverage large-scale unlabeled multi-view captures.

\paragraph{Cross-view consistency.}
We apply an explicit consistency loss on predicted albedo across overlapping views. 
For a view pair $(i,j)$ with overlap mask $M_{ij}$ and warping operator $\mathcal{W}_{i \leftarrow j}$, we define
\begin{equation}
    \mathcal{L}_{\mathrm{cv}} = \sum_{(i,j)} \big\| M_{ij} \odot 
    \big( \hat{A}_i - \mathcal{W}_{i \leftarrow j}(\hat{A}_j) \big) \big\|_1.
\end{equation}
Because $\mathcal{W}_{i \leftarrow j}$ depends on depth and camera poses, we apply this loss only to synthetic data with reliable geometry.
This encourages scene-consistent albedo while leaving shading and residual to model view- and illumination-dependent effects. 

\paragraph{Adapter and auxiliary regularization.}
To encourage the factor queries to capture distinct, non-redundant information, we apply a lightweight diversity term on the per-factor task tokens together with small L2 and floor terms on the gate $\alpha^l_k$, denoted $\mathcal{L}_{\mathrm{ada}}$. When depth or surface-normal annotations are available, we additionally supervise the auxiliary geometry predictions, $\mathcal{L}_{\mathrm{aux}} = \lambda_D \mathcal{L}_D + \lambda_N \mathcal{L}_N$, providing structural cues that improve stability for diffuse shading. Detailed formulations are provided in the supplementary.

\paragraph{Final objective.}
The overall objective is
\begin{equation}
\begin{aligned}
    \mathcal{L} =\;&
    \lambda_{\mathrm{intr}} \mathcal{L}_{\mathrm{intr}}
    + \lambda_{\mathrm{recon}} \mathcal{L}_{\mathrm{recon}}
    + \lambda_{\mathrm{cv}} \mathcal{L}_{\mathrm{cv}}  \\
    &+ \lambda_{\mathrm{ada}} \mathcal{L}_{\mathrm{ada}}
    + \lambda_{R} \mathcal{L}_{R}
    + \lambda_{\mathrm{aux}} \mathcal{L}_{\mathrm{aux}} .
\end{aligned}
\end{equation}

The residual regularizer $\mathcal{L}_{R}$ penalizes large low-frequency residuals, discouraging $R$ from absorbing diffuse appearance while allowing localized non-diffuse effects.
\rev{It is necessary because $I=A\odot S+R$ alone admits degenerate solutions (e.g., $A=I$, $S=\mathbf{1}$, $R=\mathbf{0}$); the penalties keep $R$ small and localized (ablation in Sec.~\ref{sec:ablation} and the supplementary).}
For datasets without intrinsic annotations, $\mathcal{L}_{\mathrm{intr}}$ is dropped; supervision comes from $\mathcal{L}_{\mathrm{recon}}$ and applicable regularizers, including adapter, residual, auxiliary, and cross-view terms when reliable geometry is available.


%% file: table/tab_dataset.tex
\begin{table}[h!tb]
\centering
\scriptsize
\caption{\textbf{Dataset coverage and supervision.}
MVID trains on synthetic data with intrinsic labels and unlabeled real-world sequences; $A$ / $S,R$ / Rec.\ denote albedo, shading-and-residual, and RGB-reconstruction supervision.}
\setlength{\tabcolsep}{5pt}
\renewcommand{\arraystretch}{0.95}
\begin{tabular}{lccccc}
\toprule
Dataset & \#Scenes & \#Images & $A$ & $S, R$ & Rec. \\
\midrule
Hypersim~\citeyearpar{Roberts2021Hypersim}       & 353  & 34K  & \checkmark & \checkmark & \checkmark \\
InteriorVerse~\citeyearpar{zhu2022learning}      & 4.1K & 52K  & \checkmark &            & \checkmark \\
Structured3D~\citeyearpar{zheng2020structured3d} & 3.5K & 196K & \checkmark &            & \checkmark \\
RE10K~\citeyearpar{zhou2018stereo}               & 66K  & 1.8M &            &            & \checkmark \\
ScanNet++~\citeyearpar{yeshwanth2023scannet++}   & 462  & 416K &            &            & \checkmark \\
UAVLight~\citeyearpar{uavlight}                  & 80   & 150K &            &            & \checkmark \\
UAV-Synth (Ours)                                 & 50   & 120K & \checkmark & \checkmark & \checkmark \\
\bottomrule
\end{tabular}
\label{tab:training_data}
\end{table}

%% file: table/tab_hypersim_intrinsic.tex
\begin{table*}[t]
\centering
\caption{\textbf{Intrinsic decomposition, reconstruction, and auxiliary geometry on Hypersim.}
We compare against single-view intrinsic methods, generative intrinsic methods, and feed-forward multi-view inverse rendering baselines, all evaluated without per-scene optimization.
Multi-view inverse-rendering baselines (IDArb, MAIR/MAIR++) predict parametric scene properties and are reported only on supported components.
We report PSNR$\uparrow$ and LPIPS$\downarrow$ for intrinsic factors and reconstruction, and auxiliary geometry metrics: AbsRel$\downarrow$, $\delta_1\uparrow$, normal MAE$\downarrow$, and Acc@$11.25^\circ\uparrow$.
Depth is evaluated with median scale alignment on normalized depth maps.
``--'' indicates unsupported outputs.
{\color{first}\textbf{Best}}, {\color{second}second-best}, and {\color{third}third-best} outputs are highlighted.
}
\label{tab:hypersim_joint}

\setlength{\tabcolsep}{3.3pt}
\small

\begin{tabular}{ll|cc|cc|cc|cc|cc|cc|c}
\toprule
\textbf{Category} & \textbf{Method}
& \multicolumn{2}{c|}{\textbf{Albedo}}
& \multicolumn{2}{c|}{\textbf{Shading}}
& \multicolumn{2}{c|}{\textbf{Residual}}
& \multicolumn{2}{c|}{\textbf{Recon}}
& \multicolumn{2}{c|}{\textbf{Aux. Depth}}
& \multicolumn{2}{c|}{\textbf{Aux. Normal}}
& \textbf{Time} \\
&
& PSNR$\uparrow$ & LPIPS$\downarrow$
& PSNR$\uparrow$ & LPIPS$\downarrow$
& PSNR$\uparrow$ & LPIPS$\downarrow$
& PSNR$\uparrow$ & LPIPS$\downarrow$
& AbsRel$\downarrow$ & $\delta_1\uparrow$
& MAE$\downarrow$ & Acc@$11.25^\circ\uparrow$
& s/frame$\downarrow$ \\
\midrule

\multirow{4}{*}{Single-view}
& MGNet~[\citeyear{zhu2022learning}]
& 14.49 & 0.401
& -- & --
& -- & --
& -- & --
& 0.310 & 0.528
& 17.13 & 0.544
& 1.3 \\

& Colorful~[\citeyear{careaga2024colorful}]
& 18.29 & 0.312
& {\color{third}16.10} & {\color{third}0.339}
& 21.29$^\dagger$ & 0.265$^\dagger$
& -- & --
& -- & --
& -- & --
& {\color{third}0.21} \\

& RGB-X~[\citeyear{rgbx2024}]
& {\color{third}20.80} & {\color{second}0.266}
& 12.59 & 0.341
& 20.05$^\dagger$ & 0.242$^\dagger$
& -- & --
& -- & --
& {\color{second}13.41} & {\color{second}0.713}
& 5.78 \\

& DNF~[\citeyear{zheng2025dnf}]
& 17.08 & 0.383
& -- & --
& -- & --
& -- & --
& {\color{first}\textbf{0.238}} & {\color{first}\textbf{0.691}}
& 19.39 & 0.594
& 1.09 \\

\midrule

\multirow{4}{*}{\shortstack[l]{Multi-view}}
& \rev{DiffR~[\citeyear{DiffusionRenderer}]}
& 17.29 & 0.357
& 8.40$^\ddagger$ & 0.616$^\ddagger$
& 20.35$^\dagger$ & 0.234$^\dagger$
& -- & --
& 0.346 & 0.545
& 14.85 & {\color{first}\textbf{0.734}}
& 2.03 \\

& IDArb~[\citeyear{li2024idarb}]
& 16.09 & 0.403
& -- & --
& -- & --
& -- & --
& -- & --
& 28.13 & 0.432
& 0.92 \\

& MAIR~[\citeyear{choi2023mair}]
& 15.12 & 0.412
& -- & --
& -- & --
& -- & --
& -- & --
& 19.32 & 0.461
& 8.8 \\

& MAIR++~[\citeyear{choi2025mairpp}]
& 16.98 & 0.388
& -- & --
& -- & --
& -- & --
& -- & --
& 18.32 & 0.491
& 7.3 \\

\midrule

\multirow{2}{*}{\textbf{Ours}}
& MVID-Single
& {\color{second}23.38} & {\color{third}0.270}
& {\color{second}19.41} & {\color{second}0.323}
& {\color{second}24.53} & {\color{second}0.281}
& {\color{second}24.77} & {\color{second}0.261}
& {\color{third}0.291} & {\color{third}0.546}
& {\color{third}13.96} & 0.584
& {\color{first}\textbf{0.15}} \\

& \textbf{MVID}
& {\color{first}\textbf{24.72}} & {\color{first}\textbf{0.246}}
& {\color{first}\textbf{20.11}} & {\color{first}\textbf{0.308}}
& {\color{first}\textbf{25.80}} & {\color{first}\textbf{0.264}}
& {\color{first}\textbf{25.58}} & {\color{first}\textbf{0.251}}
& {\color{second}0.266} & {\color{second}0.601}
& {\color{first}\textbf{13.30}} & {\color{third}0.600}
& {\color{second}0.19} \\

\bottomrule
\end{tabular}

{
$^\dagger$ Residual is computed as $R\!=\!I_{\text{gt}}\!-\!A\odot S$ instead of predicted; metrics are not directly comparable.
$^\ddagger$ Shading is computed from predicted diffuse and basecolor.
}
\end{table*}

%% file: section/4_experiments2.tex
\input{table/tab_hypersim_consistency}

\begin{figure*}[t]
\centering
\includegraphics[width=0.98\textwidth]{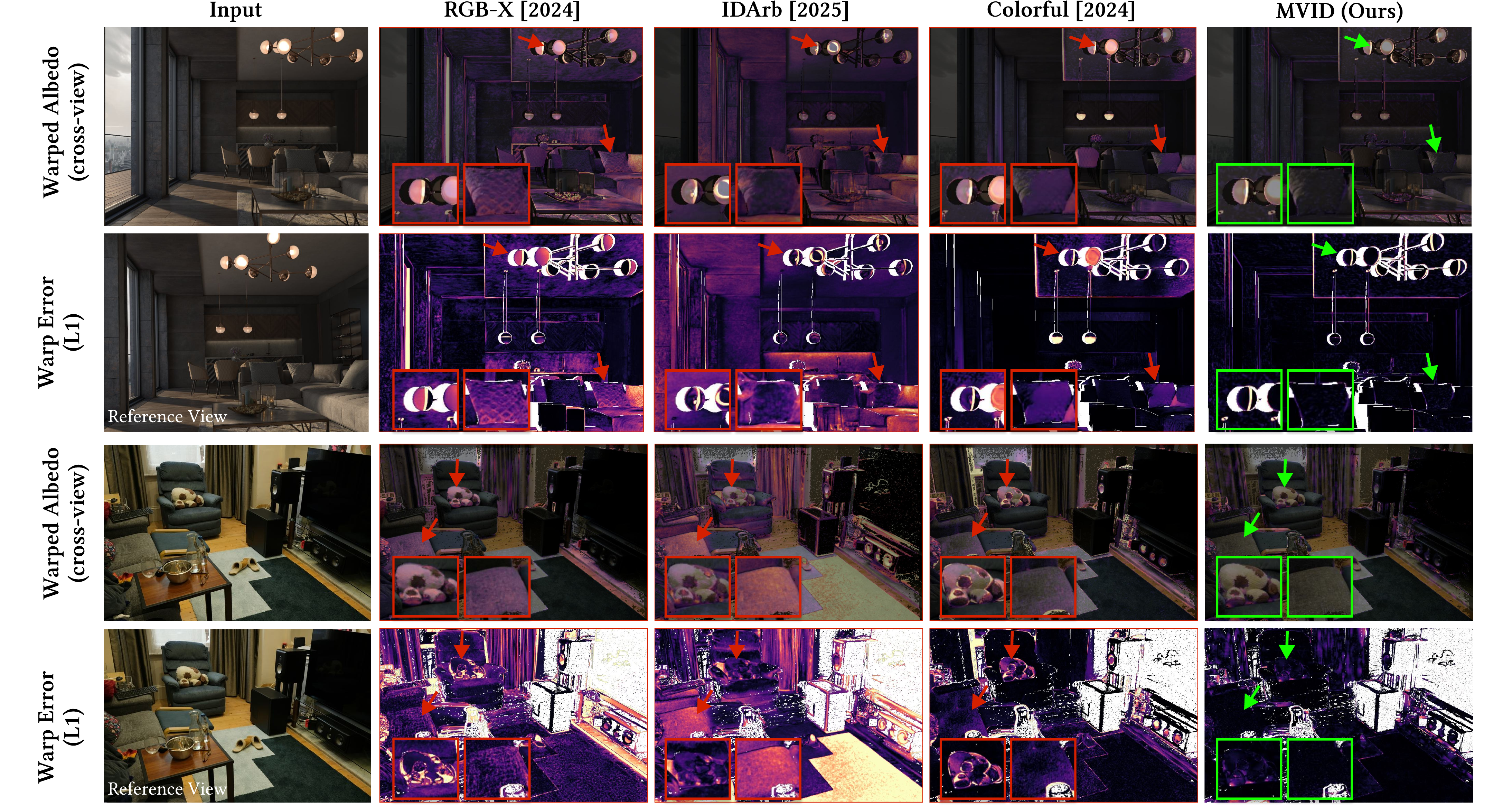}
\caption{\textbf{Cross-view albedo consistency.}
For a synthetic Hypersim scene (top two rows) and a real-world Mip-NeRF~360 Room scene (bottom two rows), source-view albedo is warped to the reference view (first column), shown with its L1 warp-error map over valid overlapping pixels (darker is more consistent).
MVID produces more stable albedo with lower warp errors, especially in the highlighted regions.}
\label{fig:consist_compare}
\end{figure*}

\section{Experiments}

We evaluate MVID under a unified feed-forward protocol without per-scene optimization.
Our experiments focus on two questions:
(i) whether MVID improves intrinsic decomposition accuracy under the residual image formation model, and
(ii) whether multi-view evidence improves view-consistent intrinsic factors.
We evaluate these properties on synthetic datasets with ground-truth intrinsic buffers and real-world multi-view sequences without intrinsic annotations.
\rev{Together, they validate our motivation at three levels: outcome (best cross-view consistency among baselines, Table~\ref{tab:albedo_cv_synth_real}), mechanism (accuracy and consistency grow from single-view to multi-view and with more input views, Tables~\ref{tab:hypersim_joint} and~\ref{tab:ablation}, Fig.~\ref{fig:view_number_ablation}), and components (removing the adapter, cross-view loss, or image-formation constraint degrades both, Table~\ref{tab:ablation}).}
Finally, we provide editing applications to demonstrate the practical utility of the predicted factors.


\subsection{Experimental Setup}
\label{sec:exp_setup}

\paragraph{Datasets.}
We use synthetic and real-world multi-view datasets covering indoor and outdoor scenes (Table~\ref{tab:training_data}).
For training, Hypersim~\cite{Roberts2021Hypersim}, InteriorVerse~\cite{zhu2022learning}, Structured3D~\cite{zheng2020structured3d}, and UAV-Synth provide intrinsic supervision, while RealEstate10K~\cite{zhou2018stereo}, ScanNet++~\cite{yeshwanth2023scannet++}, and UAVLight~\cite{uavlight} provide unlabeled RGB sequences supervised through the image formation constraint.
For evaluation, we use a fixed subset selected from the official Hypersim test split with 15 unseen scenes and 5 views per scene, and use posed scenes from Mip-NeRF~360~\cite{barron2022mipnerf360}, Deep Blending~\cite{hedman2018deep}, and Tanks\,\&\,Temples~\cite{knapitsch2017tanks} for real-world cross-view evaluation and qualitative comparison.
The supplementary provides the full dataset split, and additionally reports zero-shot measured-albedo evaluation on MAW~\cite{wu2023measured} using the official masks and evaluator.


\paragraph{Baselines.}
We evaluate MVID with representative single-view and multi-view feed-forward methods.
Single-view baselines include MGNet~\cite{zhu2022learning}, Colorful Diffuse~\cite{careaga2024colorful}, DNF-Intrinsic \cite{zheng2025dnf}, and RGB-X~\cite{rgbx2024}.
Multi-view \rev{and video-capable} baselines include IDArb~\cite{li2024idarb}, MAIR~\cite{choi2023mair}, MAIR++~\cite{choi2025mairpp}\rev{, and Diffusion Renderer~\cite{DiffusionRenderer}, which we run in multi-frame mode with the official code}.
Among these, Colorful Diffuse is our most directly comparable single-view baseline as it adopts the same residual image formation model; multi-view inverse rendering baselines (MAIR/MAIR++) target full SVBRDF and lighting recovery and are reported on the components they predict.
We additionally include \textbf{MVID-Single}, which applies the same MVID model to each view independently, isolating the contribution of multi-view aggregation.
\rev{The compared methods are not trained on identical data; a discussion of the training-data differences and why they do not undermine the comparison is provided in the supplementary.}

\paragraph{Metrics.}
For intrinsic factors and reconstruction, we report PSNR and LPIPS~\cite{zhang2018lpips, huynh2008scope}.
For cross-view consistency, we warp predicted albedo maps to a reference view using corresponding poses and depth, and report A-CV PSNR, MAE, and LPIPS over valid overlaps.
Additional MAW and cross-domain evaluation metrics are reported in the supplementary.

\begin{figure*}[t] \centering \includegraphics[width=0.99\textwidth]{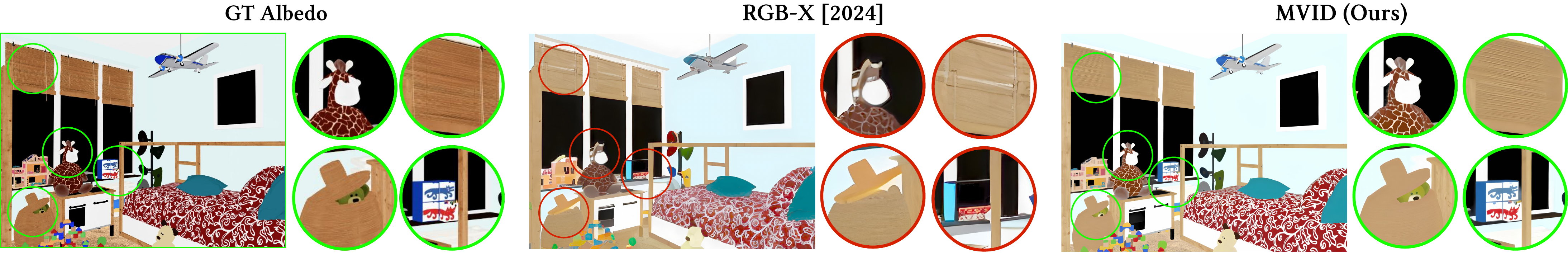}
\caption{\textbf{Sharpness vs. faithfulness.}
Single-view generative priors may favor perceptually sharp albedo details in ambiguous regions, but such details can deviate from the GT.
In contrast, MVID leverages multi-view evidence to recover albedo that is more faithful to the underlying scene materials.}
\label{fig:sharpness_vs_faithfulness}
\end{figure*}

\subsection{Per-Frame Decomposition Quality}
\label{sec:per_frame}

On the Hypersim test split, MVID achieves the best results on A/S/R decomposition and RGB reconstruction, covering albedo, shading, residual, and reconstructed RGB (Table~\ref{tab:hypersim_joint}).
It outperforms single-view intrinsic methods, generative intrinsic methods, and feed-forward multi-view inverse rendering baselines on their supported decomposition outputs.
Although geometry is only auxiliary, MVID remains competitive, ranking second on both depth metrics and first on normal MAE among comparable methods.
MVID-Single already surpasses RGB-X and Diffusion Renderer on most intrinsic metrics, while full MVID further improves decomposition, reconstruction, and geometry through shared multi-view evidence.
Qualitatively, MVID produces cleaner albedo and more structured diffuse shading than Colorful \cite{careaga2024colorful} (Fig.~\ref{fig:realworld_decomp_compare}).
Compared with existing single-view and multi-view albedo baselines, MVID also reduces cross-view albedo drift and shadow leakage (Fig.~\ref{fig:qual_albedo}).

\subsection{Cross-View Consistency}
\label{sec:cross_view}

We evaluate cross-view albedo consistency by reprojecting predicted albedo to a reference view and measuring agreement over valid overlapping regions, using 15 held-out Hypersim scenes with ground-truth geometry (5 views each) and 6 held-out real-world scenes with COLMAP-estimated~\cite{colmap} geometry (20 views each).
Table~\ref{tab:albedo_cv_synth_real} shows that image-formation decomposition methods, including Colorful Diffuse and ours, achieve stronger cross-view albedo consistency than BRDF-oriented methods, with MVID obtaining the best A-CV scores on both synthetic and real-world settings.
Fig.~\ref{fig:consist_compare} further visualizes this trend: baselines with lower A-CV scores show stronger albedo drift and brighter warp-error maps across views.
Nearby-view comparisons reveal a similar failure mode: shadows leak into albedo, albedo colors drift between frames, and diffuse shading absorbs part of the material color (see Fig.~\ref{fig:realworld_decomp_compare}).
Fig.~\ref{fig:real-world_captures} further compares multiple baselines on a real-world Kitchen \cite{barron2022mipnerf360} sequence sampled every other frame: single-view methods show view-dependent albedo drift, while diffusion-prior methods may introduce plausible but incorrect details in ambiguous regions.
MVID produces the most stable albedo across viewpoints while preserving scene structure and material appearance.
Additional video comparisons are provided in the supplementary material.

\begin{figure*}[t] \centering \includegraphics[width=0.99\textwidth]{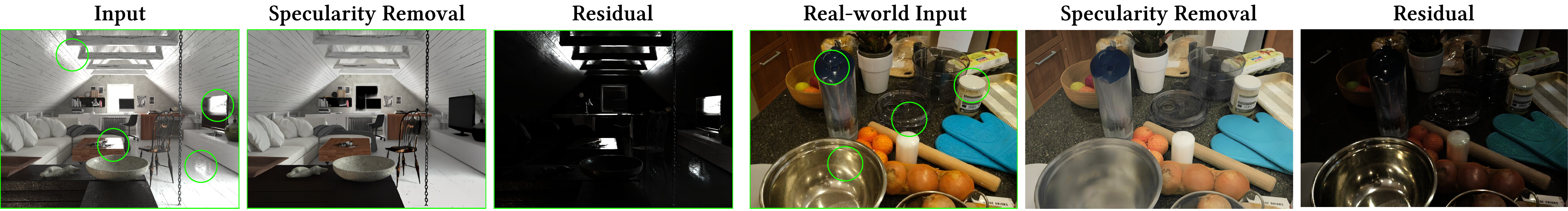}
\caption{
\textbf{Specularity removal.}
MVID decomposes synthetic and real-world inputs into a clean diffuse part and residuals.
}
\label{fig:specular_remove}
\end{figure*}

\begin{figure*}[t] \centering \includegraphics[width=0.99\textwidth]{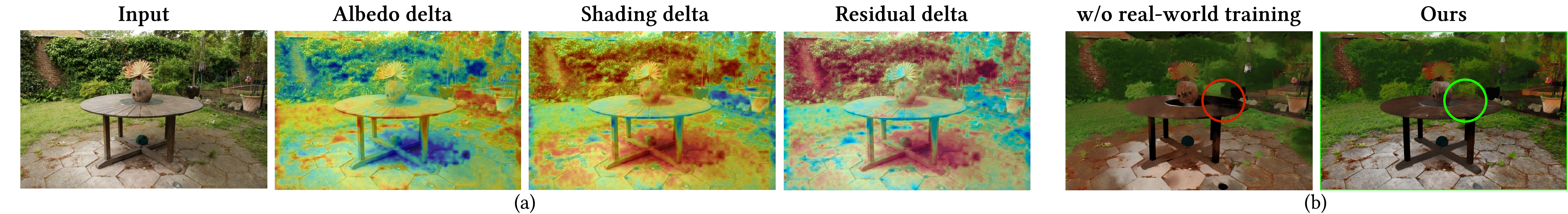}
\caption{
\textbf{Qualitative ablation results.}
(a) Adapter-induced feature deltas show emergent factor specialization without attention supervision: albedo responds to material and color cues, shading and residual to shadows and specular highlights.
(b) Without real-world image-formation supervision, shadow leakage remains in albedo; unlabeled real-world sequences yield cleaner, more stable albedo.
}
\label{fig:ablation}
\end{figure*}

\subsection{Sharpness vs.\ Faithfulness}
\label{sec:sharpness_faithfulness}
In shadowed or weakly textured regions, a single view provides limited evidence for disambiguating intrinsic factors.
Single-view diffusion methods may therefore produce visually sharp details from learned priors, but these details are not always anchored to the observed scene.
On Hypersim, this appears as competitive LPIPS but worse PSNR (Table~\ref{tab:hypersim_joint}); zoom-ins show over-sharpened structures that deviate from the ground truth and drift across viewpoints (Fig.~\ref{fig:sharpness_vs_faithfulness}).
By aggregating multi-view evidence, MVID recovers more faithful and cross-view stable intrinsics.
\rev{A representative case is the near-saturated tabletop in Fig.~\ref{fig:qual_albedo} (first row): generative priors fill this weakly-constrained region with flat, clean-looking content that is not anchored to the scene, whereas MVID recovers a smoother but view-consistent estimate closer to the ground-truth albedo (first column); the converse is that MVID can appear over-smoothed in such regions.}

\input{table/tab_ablation_intrinsic}

\begin{figure*}[t]
\centering
\includegraphics[width=0.95\textwidth]{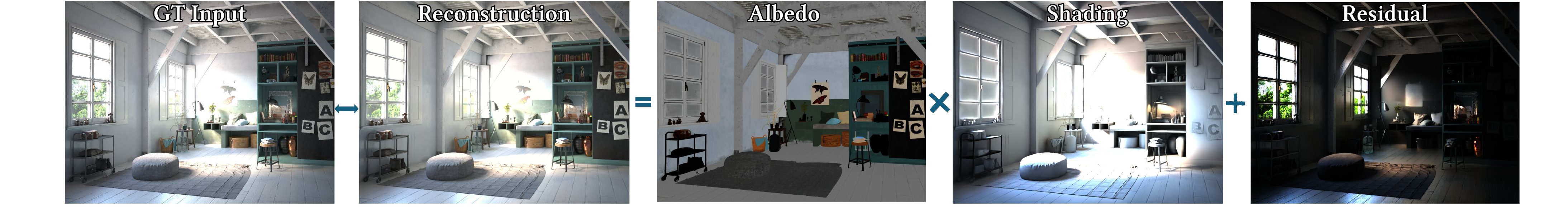}
\caption{
\textbf{Residual image formation and multi-view consistency.}
MVID reconstructs each input as $\hat{\mathbf{I}}=\hat{\mathbf{A}}\odot\hat{\mathbf{S}}+\hat{\mathbf{R}}$; the residual captures view-dependent and non-diffuse appearance.
}
\label{fig:reconstruction}
\end{figure*}

\subsection{Downstream Applications}
\label{sec:applications}

\paragraph{Multi-view consistent illumination editing.}
We build an interactive editing tool (detailed in the supplementary) for image-domain illumination editing: by modifying $\hat{S}$ and $\hat{R}$ while keeping $\hat{A}$ fixed, users can adjust illumination appearance, color temperature, and shading direction.
Because MVID produces stable albedo across views, the same edit propagates consistently across viewpoints (Fig.~\ref{fig:teaser}; more examples in the supplementary); methods that produce view-inconsistent factors yield visible artifacts when the same edit is applied across multiple views.
\rev{Editing is a downstream, image-space application rather than a primary contribution or physically based relighting; following prior intrinsic methods~\cite{careaga2024colorful, rgbx2024, zheng2025dnf}, we validate it qualitatively and leave multi-illumination relighting evaluation to a physical-relighting extension.}

\paragraph{Specularity removal.}
Following~\cite{careaga2024colorful}, the diffuse image $\hat{A} \odot \hat{S}$ provides a specularity-free version of the input.
Fig.~\ref{fig:specular_remove} shows specularity removal results on real-world scenes, where MVID effectively suppresses highlights while preserving underlying surface texture; multi-view consistency further ensures that the removed regions remain stable across views.

\subsection{Ablation Study}
\label{sec:ablation}

\paragraph{Multi-view evidence.}
MVID improves over MVID-Single in albedo, shading, and A-CV (Tables~\ref{tab:hypersim_joint},~\ref{tab:albedo_cv_synth_real},~\ref{tab:ablation}), confirming the value of scene-level aggregation; on Hypersim, these gains increase with more input views and saturate after the default 5-view setting (Fig.~\ref{fig:view_number_ablation}).

\input{table/tab_pgfplots}

\paragraph{Factor query adapter.}
Removing the multi-layer factor query adapter degrades per-factor accuracy and cross-view consistency (Table~\ref{tab:ablation}).
Fig.~\ref{fig:ablation}(a) shows emergent specialization in the adapter-induced feature delta: albedo attends to material/color cues, while shading and residual focus on shadows and specular highlights.

\rev{\paragraph{Local refinement for albedo.}
Retraining without the gated ResNeXt local-detail fusion in the albedo head reduces albedo PSNR from 24.72 to 24.01 while leaving cross-view consistency essentially unchanged (A-CV 25.12), and visibly softens high-frequency texture (visual comparison in the supplementary): the CNN branch contributes pixel-aligned detail, while the multi-view tokens remain responsible for consistency.}

\rev{\paragraph{Residual regularizer.}
Removing $\mathcal{L}_R$ lets the residual grow into a free reconstruction channel that absorbs diffuse appearance; the visual ablation in the supplementary shows this degeneration, and that $R$ stays small and localized under the full model.}

\paragraph{Cross-view consistency loss.}
Removing $\mathcal{L}_{\mathrm{cv}}$ markedly lowers cross-view albedo consistency while leaving per-frame accuracy largely intact (Table~\ref{tab:ablation}), showing that the explicit term complements the implicit coupling from the shared representation.

\paragraph{Image formation constraint and real-world training.}
Removing $\mathcal{L}_{\mathrm{recon}}$ causes the largest drop in per-factor accuracy and reduces cross-view consistency (Table~\ref{tab:ablation}), confirming the need for a reconstruction anchor that couples albedo, shading, and residual (Fig.~\ref{fig:reconstruction}).
Without real-world data, the model generalizes poorly in the wild, leaving stronger shadow leakage in albedo (Fig.~\ref{fig:ablation} b) and lower real-world A-CV despite similar synthetic PSNR.

\paragraph{Auxiliary geometry supervision.}
Depth and normal supervision encourage the shared multi-view representation to encode scene structure.
Removing it moderately hurts diffuse shading, with smaller effects on albedo and cross-view consistency (Table~\ref{tab:ablation}).

%% file: table/tab_hypersim_consistency.tex
\begin{table}[t]
\centering
\caption{\textbf{Cross-view albedo consistency on synthetic and real-world scenes.}
Warped-albedo agreement over overlapping views; Hypersim uses ground-truth geometry, real-world scenes use estimated poses and depth.
SV/MV denote single-view and multi-view (or video-capable) methods.}
\label{tab:albedo_cv_synth_real}

\scriptsize
\setlength{\tabcolsep}{3pt}
\renewcommand{\arraystretch}{0.88}

\resizebox{0.98\columnwidth}{!}{%
\begin{tabular}{c l ccc ccc}
\toprule

\multirow{2}{*}{\textbf{Type}}
& \multirow{2}{*}{\textbf{Method}}

& \multicolumn{3}{c}{\textbf{Hypersim}}
& \multicolumn{3}{c}{\textbf{Real-world}} \\

\cmidrule(lr){3-5}
\cmidrule(lr){6-8}

&
& PSNR$\uparrow$
& LPIPS$\downarrow$
& MAE$\downarrow$

& PSNR$\uparrow$
& LPIPS$\downarrow$
& MAE$\downarrow$ \\

\midrule

\multirow{3}{*}{\rotatebox[origin=c]{0}{\scriptsize SV}}

& RGB-X~\citeyearpar{rgbx2024}
& 19.80 & 0.186 & 0.060
& 19.07 & 0.167 & 0.073 \\

& Colorful~\citeyearpar{careaga2024colorful}
& 25.22 & 0.129 & 0.033
& 24.15 & 0.093 & 0.041 \\

& \textbf{MVID-Single}
& 25.51 & 0.091 & 0.027
& 24.21 & 0.078 & 0.044 \\

\midrule

\multirow{3}{*}{\rotatebox[origin=c]{0}{\scriptsize MV}}

& \rev{DiffR~\citeyearpar{DiffusionRenderer}}
& 20.65 & 0.164 & 0.056
& 22.60 & 0.106 & 0.053\\

& IDArb~\citeyearpar{li2024idarb}
& 18.17 & 0.224 & 0.105
& 18.52 & 0.220 & 0.092 \\

& \textbf{MVID (Ours)}
& \textbf{26.68} & \textbf{0.083} & \textbf{0.020}
& \textbf{25.14} & \textbf{0.064} & \textbf{0.035} \\

\bottomrule
\end{tabular}
}

\end{table}

%% file: table/tab_ablation_intrinsic.tex
\begin{table}[h!tb]
\centering
\scriptsize
\caption{\textbf{Ablation study.}
We report albedo/shading PSNR on Hypersim and real-world cross-view consistency (A-CV).}
\setlength{\tabcolsep}{5.5pt}
\renewcommand{\arraystretch}{0.95}
\begin{tabular}{lccc}
\toprule
\textbf{Variant}
& $A$ PSNR$\uparrow$
& $S$ PSNR$\uparrow$
& A-CV PSNR$\uparrow$ \\
\midrule
\textbf{MVID}
& \textbf{24.72}
& \textbf{20.11}
& \textbf{25.14} \\
\cmidrule(lr){1-4}
MVID-Single
& 23.38
& 19.41
& 24.21 \\
w/o factor query adapter
& 22.85
& 18.74
& 23.62 \\
w/o cross-view consistency loss
& 24.38
& 19.92
& 22.95 \\
w/o image formation constraint
& 22.80
& 18.60
& 23.18 \\
w/o real-world RGB training
& 24.45
& 19.95
& 22.70 \\
w/o auxiliary geometry supervision
& 24.18
& 18.95
& 24.55 \\
\rev{w/o local-detail fusion (albedo CNN)}
& \rev{24.01}
& \rev{20.10}
& \rev{25.12} \\

\bottomrule
\end{tabular}
\label{tab:ablation}
\end{table}

%% file: table/tab_pgfplots.tex
\begin{figure}[t]
\centering
\begin{tikzpicture}
\begin{axis}[
    width=0.74\linewidth,
    height=0.42\linewidth,
    xmin=1, xmax=8,
    xtick={1,2,3,5,8},
    grid=major,
    legend style={
        at={(1.02,0.5)},
        anchor=west,
        legend columns=1,
        font=\tiny,
        draw=none,
        fill=none,
        row sep=-1pt
    },
    tick label style={font=\tiny},
    label style={font=\scriptsize},
    line width=0.7pt,
    mark size=1.5pt,
]
\addplot+[mark=o] coordinates {
    (1,23.38) (2,24.10) (3,24.55) (5,24.72) (8,24.78)
};
\addlegendentry{$A$ PSNR}

\addplot+[mark=square*] coordinates {
    (1,19.41) (2,19.78) (3,20.00) (5,20.11) (8,20.15)
};
\addlegendentry{$S$ PSNR}

\addplot+[mark=triangle*] coordinates {
    (1,25.51) (2,26.05) (3,26.42) (5,26.68) (8,26.72)
};
\addlegendentry{A-CV}

\end{axis}
\end{tikzpicture}
\caption{\textbf{Effect of input view number.}
Hypersim albedo PSNR, shading PSNR, and A-CV vs.\ the number of input views.
All three improve monotonically, with the largest gain from moving beyond a single view; improvements saturate near the default 5-view setting.}
\label{fig:view_number_ablation}
\end{figure}

%% file: section/5_conclusion.tex
\section{Discussion and Conclusion}
\label{sec:limitations}
We presented \textbf{MVID}, a feed-forward approach for multi-view intrinsic image decomposition under the residual image formation model.
By aggregating sparse overlapping views into a shared scene representation, MVID predicts \rev{a view-consistent albedo together with coherent per-view diffuse shading and residual factors}.
Our experiments show that multi-view evidence improves both decomposition quality and cross-view consistency, and the resulting intrinsics support \rev{image-space} applications such as illumination editing and specularity removal.

\rev{\paragraph{Limitations.}
Thin structures can be distorted (e.g., the stair railing in Fig.~\ref{fig:teaser}), reflecting the aggregator's patch-token resolution.
In under-constrained regions, MVID trades hallucinated sharpness for faithfulness and can look over-smoothed (Sec.~\ref{sec:sharpness_faithfulness}).
Real-world generalization remains incomplete: dark, transparent, reflective, and strongly view-dependent regions are difficult (failure cases in the supplementary), and some edits retain a synthetic look.
Finally, the residual is an image-domain correction rather than a physical factor, so our edits are image-space manipulations.}

\paragraph{\rev{Future work.}}
MVID assumes static scenes with cross-view coverage; dynamic in-the-wild captures remain challenging, as moving objects and time-varying illumination violate multi-view consistency. The implementation also fixes the number of input views per forward pass due to GPU memory; streaming or memory-efficient cross-view attention could extend MVID to dense captures and long videos.
Beyond scalability, physically factorizing the residual into editable components such as specularity and transparency, and coupling the predicted factors with explicit lighting estimation, would move image-space editing toward physically based relighting.
Integrating MVID with explicit 3D representations for jointly consistent view synthesis and relighting is another promising direction.